\pdfoutput=1

\documentclass[11pt]{article}

\usepackage[final]{acl}

\usepackage{times}
\usepackage{latexsym}

\usepackage[T1]{fontenc}

\usepackage[utf8]{inputenc}

\usepackage{microtype}

\usepackage{inconsolata}

\usepackage{graphicx}

\usepackage{linguex}
\usepackage{multirow}
\usepackage{booktabs}
\usepackage{array}
\usepackage{pifont} 
\usepackage{url}
\usepackage{subfig}
\usepackage{amsmath}

\usepackage{caption}
\usepackage{todonotes}
\usepackage{paralist}
\usepackage{algorithm}
\usepackage{algpseudocode}
\usepackage{mathtools}
\usepackage{etoolbox}
\usepackage{amsfonts}
\usepackage{seqsplit}

\usepackage{CJKutf8}

\title{A Structured Framework for Evaluating and Enhancing Interpretive Capabilities of Multimodal LLMs in Culturally Situated Tasks}

\author{
  \textbf{Haorui Yu\textsuperscript{1}},
  \textbf{Ramon Ruiz-Dolz\textsuperscript{2}}, and
  \textbf{Qiufeng Yi\textsuperscript{3}}
\\
  \textsuperscript{1}DJCAD, University of Dundee, United Kingdom \\
  \textsuperscript{2}ARG-tech, SSEN, University of Dundee, United Kingdom \\
  \textsuperscript{3}School of Computer Science, University of Birmingham, United Kingdom\\
  \texttt{\{2655435, rruizdolz001\}@dundee.ac.uk} \\
  \texttt{qxy953@student.bham.ac.uk}
}

\begin{document}
\begin{CJK}{UTF8}{gbsn}
\maketitle

\begin{abstract}
This study aims to test and evaluate the capabilities and characteristics of current mainstream Visual Language Models (VLMs) in generating critiques for traditional Chinese painting. To achieve this, we first developed a quantitative framework for Chinese painting critique. This framework was constructed by extracting multi-dimensional evaluative features covering evaluative stance, feature focus, and commentary quality from human expert critiques using a zero-shot classification model. Based on these features, several representative critic personas were defined and quantified. This framework was then employed to evaluate selected VLMs such as Llama, Qwen, or Gemini. The experimental design involved persona-guided prompting to assess the VLM's ability to generate critiques from diverse perspectives. Our findings reveal the current performance levels, strengths, and areas for improvement of VLMs in the domain of art critique, offering insights into their potential and limitations in complex semantic understanding and content generation tasks. The code used for our experiments can be publicly accessed at: \url{https://github.com/yha9806/VULCA-EMNLP2025}.
\end{abstract}

\section{Introduction}


Large language models (LLMs) have demonstrated remarkable performance on general NLP benchmarks, yet their applicability in culturally embedded, humanistic domains remains limited. In high-context interpretive tasks such as art criticism, clinical narrative analysis, or historical commentary, model performance depends not only on linguistic fluency or factual accuracy, but also on deeper forms of cognitive alignment—epistemic sensitivity, rhetorical coherence, and cultural adaptability.

A representative and particularly demanding testbed for such capabilities is \textit{Chinese art commentary}. This genre, especially when analyzing works like traditional landscape or court paintings, involves symbolic interpretation, aesthetic judgment, and deeply situated cultural discourse. Existing multimodal LLMs are rarely evaluated in this space. Standard benchmarks such as MME~\cite{fu2025mme} and MMBench~\cite{liu2024mmbench} focus on object recognition or task-oriented vision-language reasoning, while frameworks like ArtGPT~\cite{yuan2024artgpt4artisticunderstandinglargevisionlanguage} emphasize captioning and factual grounding. These methods largely overlook interpretive nuance and disciplinary diversity.

Meanwhile, humanistic commentary often exhibits non-linear logic, specialized lexicons, and varied stylistic conventions, particularly in Chinese art contexts where rhetorical strategies such as \textit{yijing} (意境, artistic conception) or \textit{qiyun shengdong} (气韵生动, spiritual resonance) are essential but difficult to quantify~\cite{bush1971chinese, siren1936chinese}. Without appropriate grounding, LLMs risk producing synthetic outputs that mimic surface patterns but fail to demonstrate epistemic alignment~\cite{guo2023evaluating}. This growing mismatch calls for new paradigms in evaluation and adaptation.

To address these challenges, we introduce VULCA—the \textit{Vision-Understanding and Language-based Cultural Adaptability Framework}. VULCA is a structured evaluation and enhancement framework designed to assess how well VLMs align with domain-specific interpretive practices in culturally situated tasks. Our work centers on Chinese art commentary, but the methodology generalizes to other multimodal and epistemically rich domains such as religion, medicine, or history. VULCA combines three core components: (1) a multi-dimensional human expert benchmark (MHEB) constructed from 163 art commentaries annotated across five cultural capability dimensions; (2) a persona-guided recontextualization mechanism using eight interpretive personas and a domain-specific knowledge base; and (3) a joint evaluation pipeline integrating vector-space semantic alignment with rubric-based capability scoring. Commentaries are generated from annotated traditional Chinese paintings, and their alignment with expert patterns is evaluated with and without interventions. As a result, we produce five contributions: (i) the definition of VULCA, a new structured framework for assessing and enhancing VLMs in culturally grounded, multimodal reasoning tasks; (ii) we construct MHEB, a high-quality human benchmark of Chinese art commentary annotated across five capability dimensions; (iii) we develop and evaluate persona-guided recontextualization interventions using eight expert personas and a domain-specific knowledge base; (iv) we demonstrate over 20\% improvement in symbolic reasoning and over 30\% improvement in argumentative coherence on Gemini 2.5 Pro using our proposed method; and (v) we establish the generalizability of our evaluation methodology to other epistemically rich domains such as religion, history, and education.

Together, our work highlights the need for new evaluation paradigms that go beyond benchmark metrics and toward measuring how well LLMs can adapt to the interpretive demands of real-world, interdisciplinary contexts.

\section{Related Work}

\paragraph{Missing Evaluation Dimensions for Cultural Reasoning.}
Despite significant advances in multimodal evaluation, current benchmarks primarily target factual understanding rather than cultural interpretation. Existing benchmarks for large or multimodal language models, such as \cite{fu2025mme, you2024ferret}, emphasize factual accuracy or instruction following, seldom addressing symbolic interpretation or epistemic alignment. Recent cultural evaluation efforts like M3Exam \cite{zhang_m3exam_2023} and SEED-Bench \cite{Li_2024_CVPR}, although primarily focused on multimodal understanding with limited cultural coverage, begin to incorporate cultural knowledge but focus on factual recall rather than interpretive reasoning. ArtGPT \cite{yuan2024artgpt4artisticunderstandinglargevisionlanguage}, for instance, evaluates stylistic generation but lacks formal metrics for interpretive depth. While prior work explores aesthetic reasoning \cite{wang_changes_2024}, these studies rarely offer structured, multi-capability evaluation. Our work addresses this gap by introducing cultural adaptability, operationalized through a multi-dimensional human expert benchmark with capability rubrics, enabling quantitative comparison in high-context domains like Chinese art.

\paragraph{Limitations of Persona Conditioning Without Grounding.}
Building on evaluation gaps, current persona-based approaches show promise but remain limited in cultural domains. Persona use in LLM evaluation shows promise for style control \cite{jiang-etal-2024-personallm, wang-etal-2024-rolellm}, yet most methods lack structured knowledge grounding, especially in epistemically rich domains. While recent work on role-playing \cite{shanahan_role_play_2023} and character conditioning demonstrates behavioral adaptation, these approaches often rely on surface-level stylistic changes rather than deep domain expertise. Our method addresses this limitation by combining persona simulation with curated domain-specific knowledge to guide generation towards symbolic reasoning and cultural interpretation, not just stylistic alignment, offering a controlled intervention mechanism.

\paragraph{Gap in Multimodal Input–Interpretation Evaluation.}
Current multimodal frameworks like MMBench or LLaVA \cite{liu_visual_2023} primarily focus on classification, question answering, or instruction following, rarely requiring grounded interpretation. Our pipeline links annotated symbolic elements with structured prompts for art commentary, evaluating VLM outputs for semantic alignment with MHEB using vector-space and rubric-based metrics, addressing a gap in assessing image-conditioned cultural reasoning.

\paragraph{Lack of Comparative Cultural Interventions Across Models.}
Surveys \cite{guo2023evaluating} discuss LLM limitations in nuanced discourse, but few studies compare model responsiveness to structured cultural interventions. Our empirical evaluation shows persona and knowledge base intervention improves symbolic reasoning and argumentative coherence by over 20–30\%, highlighting epistemic alignment's role beyond fluency. This cross-model, capability-specific analysis distinguishes our work.

\section{Methodology}

\begin{figure}[htbp] 
    \centering
    \includegraphics[width=\linewidth, keepaspectratio]{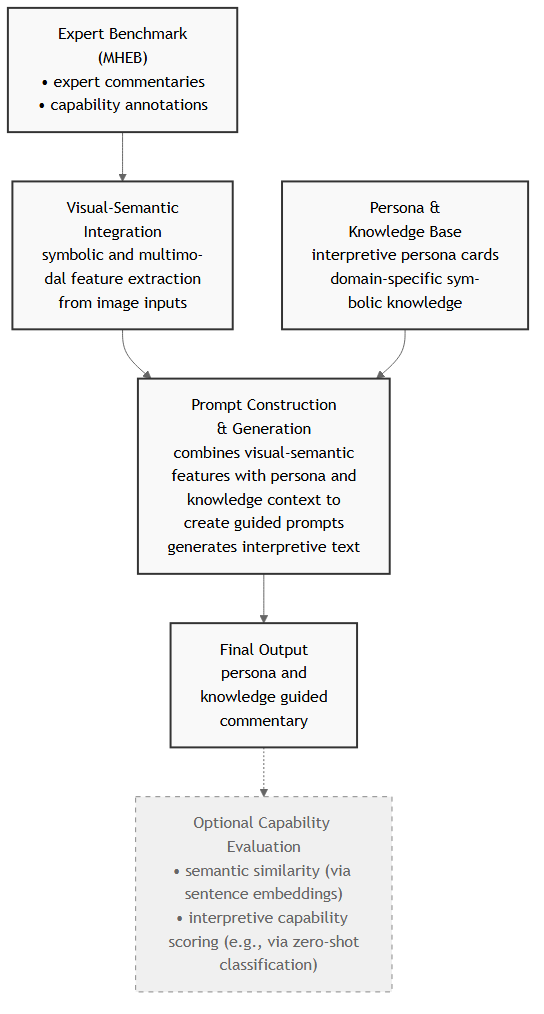} 
    \caption{Overview of the VULCA framework, illustrating its components and their interactions for structured evaluation and intervention in art criticism.}
    \label{fig:vulca_framework}
\end{figure}

This research aims to comprehensively evaluate Visual Language Models (VLMs) capabilities in generating critiques for traditional Chinese painting, assessing their understanding of image content, commentary quality, and adaptability to guided perspectives. The workflow involves: Framework Construction, developing a quantitative analytical framework from human expert commentaries, including defining evaluative dimensions and critic personas; VLM evaluation experiment design, creating structured protocols for VLM critique generation under conditions like persona-based and baseline prompting; and experimentation and result analysis, implementing experiments, collecting VLM critiques, and analyzing them with the developed framework to assess capabilities and intervention impacts. Figure~\ref{fig:vulca_framework} provides an overview of this framework and its components.

A cornerstone is the quantitative framework benchmark for VLM critiques, built upon human expert commentaries on Chinese art. To ensure objective, reproducible, and fine-grained evaluation, an automated capability assessment framework was developed. This involves feature extraction, multi-dimensional capability scoring, profile assignment, and visualization, using a zero-shot classification model for fine-grained evaluative labels. The scoring covers painting element recognition, Chinese painting understanding, and language usage, each with a dedicated rubric. This structured, rule-based approach enhances objectivity and facilitates large-scale benchmarking \cite{jiang_multimodal_2025, hayashi-etal-2025-irr}.

\subsection{MHEB Construction and Annotation Process}
\label{sec:mheb-construction}

Our three-dimensional evaluation framework synthesizes Eastern and Western art criticism traditions with modern museum documentation standards into the three major dimensions of Evaluative Stance, Feature Focus, and Commentary Quality. The framework draws from:

(1) \textit{Chinese Art Theory}: Building on Xie He's Six Canons (六法, 550 CE)~\cite{xie_he_six_canons}, particularly the concepts of ``spirit resonance'' (气韵生动) and ``bone method'' (骨法用笔), which inform our Feature Focus dimension's emphasis on brushwork technique, artistic conception, and emotional expression.

(2) \textit{Western Art Historical Methods}: Incorporating Baxandall's ``inferential criticism''~\cite{baxandall1985patterns} and Gombrich's psychological approach~\cite{gombrich1960art}, which contribute to our Evaluative Stance dimension through categories like comparative analysis, theoretical construction, and critical inquiry.

(3) \textit{Museum Documentation Standards}: Following international cataloging frameworks from ICOM-CIDOC~\cite{icom_cidoc_standards} and practices from the Palace Museum Beijing, National Palace Museum Taipei, and Metropolitan Museum of Art~\cite{met_museum_standards}, which standardize descriptive categories for artwork documentation. These inform our systematic approach to feature extraction and the structured nature of our Commentary Quality dimension.

This synthesis creates a culturally-informed yet methodologically rigorous framework that captures both the technical aspects emphasized in Western criticism (e.g., composition, color theory) and the philosophical-spiritual dimensions central to Chinese art evaluation (e.g., artistic conception, symbolic meaning). 
The MHEB was therefore systematically constructed through the following process:

\textbf{Data Collection.} We collected 163 expert commentaries from authoritative sources including museum catalogs from the Palace Museum Beijing, National Palace Museum Taipei, and Metropolitan Museum of Art, as well as peer-reviewed art history journals and monographs by recognized scholars specializing in Qing court painting. Each commentary averages 500-800 Chinese characters and provides in-depth analysis of specific paintings from the ``Twelve Months'' series. The annotation process generated 558 total annotation instances (163 texts × 3 annotators plus quality control samples), which were consolidated into 163 final records after resolving disagreements.

\textbf{Expert Sources.} The 163 commentaries in MHEB were extracted from scholarly publications by 9 distinguished art historians specializing in Chinese painting and Qing court art. The corpus includes: Xue Yongnian (薛永年, 17 texts from two monographs), Wang Di (汪涤, 28 texts), Yang Danxia (杨丹霞, 28 texts), Nie Chongzheng (聂崇正, 15 texts), Shan Guoqiang (单国强, 18 texts), Li Shi (李湜, 17 texts), Xu Jianrong (徐建融, 17 texts), Zhu Wanzhang (朱万章, 11 texts), and Chen Yunru (陈韵如, 12 texts). These experts represent major institutions including the Palace Museum Beijing, National Palace Museum Taipei, and leading Chinese art history departments, ensuring diverse yet authoritative perspectives on Giuseppe Castiglione's ``Twelve Months'' series.

\textbf{Annotation Process.} Three annotators with graduate-level training in Chinese art history independently labeled each commentary. Annotators were provided with a 20-page annotation guideline detailing the three evaluation dimensions (Evaluative Stance, Feature Focus, Commentary Quality) and their respective sub-categories. Each annotator spent approximately 15-20 minutes per commentary, assigning scores for all 38 primary feature labels using a 0-1 continuous scale based on presence and prominence, from which 9 additional analytical dimensions were derived. Annotation was performed independently using a custom web-based interface, with randomized presentation order to minimize bias.

\textbf{Quality Control Measures.} To ensure annotation quality throughout the process, we implemented multiple control mechanisms: (1) 20\% of commentaries were double-annotated to monitor consistency; (2) bi-weekly calibration sessions were held over the 3-month annotation period where annotators discussed challenging cases and aligned their understanding; (3) continuous monitoring tracked annotator performance and drift. These measures ensured that the annotation process remained consistent and reliable throughout the data collection period.

\textbf{Inter-Annotator Agreement (IAA).} To quantitatively assess the reliability of our annotations, we calculated inter-annotator agreement using two complementary metrics. For categorical labels (e.g., stance categories), we computed Fleiss' kappa~\cite{fleiss1971measuring}, which measures agreement beyond chance for multiple raters. For continuous scores (e.g., feature prominence ratings from 0-1), we calculated the intraclass correlation coefficient (ICC)~\cite{shrout1979intraclass}, which assesses the consistency of quantitative measurements across raters. The average Fleiss' kappa across stance categories was 0.78, indicating substantial agreement according to Landis and Koch's interpretation scale. The ICC for feature prominence scores reached 0.82, demonstrating excellent reliability. When disagreements occurred (defined as $\kappa$ $<$ 0.6 for specific labels), they were resolved through discussion, with a senior art historian serving as arbiter for persistent conflicts. The stable inter-rater agreement ($\kappa$ variation $<$ 0.05 across time) validated the effectiveness of our quality control measures. Final dataset statistics show balanced representation across different evaluative stances (Historical: 31\%, Aesthetic: 28\%, Technical: 23\%, Comparative: 18\%) and comprehensive coverage of feature focus.

\subsection{Feature Engineering from Human Expert Critiques}

\begin{figure*}[]
    \centering
    \includegraphics[width=0.7\textwidth, keepaspectratio]{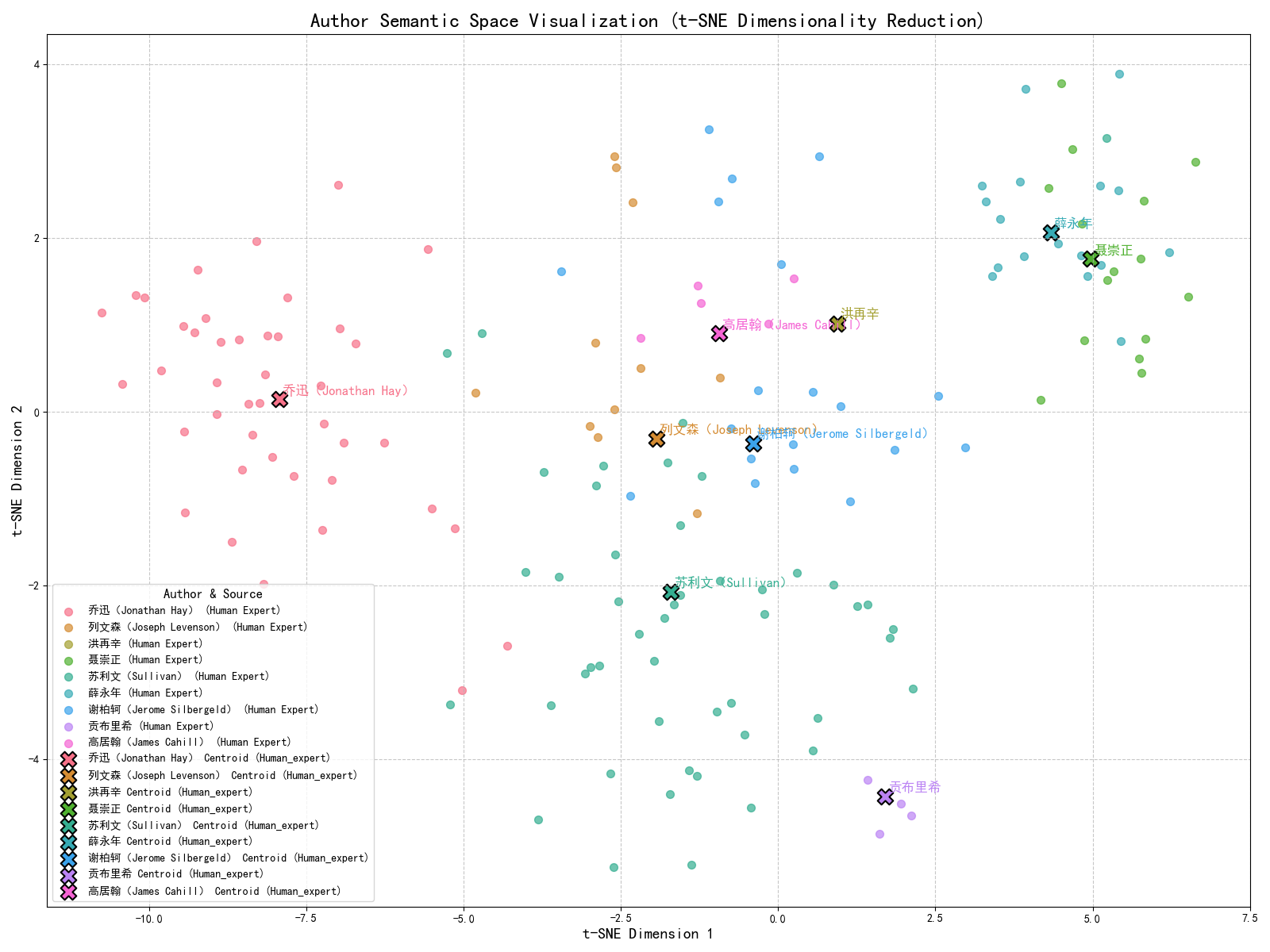}
    \caption{T-SNE visual representation of human expert art commentaries.}
    \label{fig:semantic-visual}
\end{figure*}

Framework foundation relies on human expert commentaries, significantly from Giuseppe Castiglione's (Lang Shining, 郎世宁) ``Twelve Months'' (十二月令图) series—Qing imperial court paintings fusing Chinese and Western traditions. To enhance model training and evaluation, a sliding window cropping strategy (640×640 pixel sub-images) was applied to these high-resolution images, augmenting data diversity and granularity for improved VLM detail recognition and evaluation accuracy, a common practice in computer vision (e.g., \cite{Lin2014MicrosoftCC, krishna2017visual}).

We employed a zero-shot classification model to systematically extract evaluative characteristics. Specifically, we used the multilingual BART-large-mnli model~\cite{lewis2020bart, williams2018broad}, which has been fine-tuned on natural language inference tasks and can classify text into arbitrary categories without task-specific training. For each commentary text, the model computes the probability of belonging to each predefined label using the entailment paradigm. Given a text $T$ and a label $L$, the model evaluates the hypothesis ``This text is about $L$'' and outputs a softmax probability score $p(L|T) \in [0,1]$. We apply this process across 38 labels spanning three dimensions: \textit{Evaluative Stance} (10 labels, e.g., ``Historical Research'', $p=0.85$), \textit{Feature Focus} (17 labels, e.g., ``Use of Color'', $p=0.72$), and \textit{Commentary Quality} (11 labels, e.g., ``Profound Insight'', $p=0.68$). Furthermore, we complemented this set of 38 labels with 9 additional labels representing higher level features: 5 profile alignment scores derived from clustering analysis of the 38 primary features, and 4 supplementary analytical dimensions for enhanced discrimination between critique styles.

Thresholds for binary classification were empirically determined through validation on a held-out subset: labels with $p > 0.5$ are considered present, while prominence levels are captured by the continuous scores. This comprehensive 47-dimensional feature vector (38 primary features plus 9 derived dimensions) enables nuanced quantitative comparison and clustering. Appendix~\ref{app:zeroshot_labels} provides complete list of all 47 dimensions: the 38 primary labels and 9 derived analytical dimensions. Figure~\ref{fig:semantic-visual} visualizes the MHEB semantic distribution from these features.

The zero-shot classification model serves as an analytical tool for deconstructing expert texts and building our evaluation framework, distinct from the VLMs (e.g., Gemini 2.5 Pro, Qwen-VL) evaluated later.

\subsection{Evaluation of Dimensions and Label System}

The three dimensions of our framework (i.e., Evaluative Stance, Feature Focus, and Commentary Quality) were derived from multiple sources: (1) traditional Chinese painting theory, particularly Xie He's ``Six Principles of Painting'' (谢赫六法)~\cite{acker1954some} which emphasizes spirit resonance (气韵生动), bone method (骨法用笔), and correspondence to nature (应物象形); (2) Western art criticism frameworks from Panofsky's three levels of meaning~\cite{panofsky1955meaning} and Wölfflin's formal analysis principles~\cite{wolfflin1950principles}; (3) consultations with curators from the Palace Museum and Metropolitan Museum who validated the relevance of these dimensions for Qing court painting analysis; and (4) empirical analysis of recurring patterns in our collected expert commentaries. 

Evaluative Stance characterizes the rhetorical or evaluative position taken by the commentator (e.g., historical interpretation, praise, or critique). Feature Focus identifies the specific visual or contextual aspects discussed in the commentary (e.g., line quality, symbolism, spatial composition). Commentary Quality captures the analytical depth and logical structure of the commentary, ranging from clear, well-argued insights to superficial or biased remarks. Furthermore, each dimension comprises a set of fine-grained subcategories with bilingual English–Chinese mappings. Full definitions and label lists are provided in Appendix~\ref{app:label_definitions}.




\subsection{Construction and Definition of Critic Personas}
To capture holistic critique style and depth beyond granular features, we constructed ``critic personas'' representing archetypal critical perspectives. Their development was data-driven, analyzing features from human expert commentaries, complemented by art history domain expertise. Five core personas were defined: Comprehensive Analyst (博学通论型), Historically Focused Critic (历史考据型), Technique \& Style Focused Critic (技艺风格型), Theory \& Comparison Focused Critic (理论比较型), and General Descriptive Profile (泛化描述型). These five core personas represent data-driven evaluation categories derived from clustering analysis of human expert features, serving as benchmarks for assessing whether VLM outputs align with recognizable expert critique patterns.

Each persona is quantitatively defined by rules and thresholds based on zero-shot classification feature scores. This rule-based matching objectively assigns commentaries (human or VLM) to personas. Persona definition and matching rely on explicit features and rule-based logic, not primarily direct semantic embedding of raw text. Dimensionality reduction (t-SNE/UMAP) visualizes commentary and persona distribution in the feature space, not for initial persona vector generation. 


\subsection{Task Definition}
\label{sec:exp-protocol} 

This quantitative framework guided experiments evaluating selected VLMs (e.g., Gemini 2.5 Pro, Qwen-VL). The core task required VLMs to generate commentary on provided traditional Chinese painting images. Experiments typically involved structured, multi-round interactions for each VLM per image, including persona-based and baseline Q\&A rounds.

Inputs were multifaceted: (i) high-definition ``Monthly Images'' (sometimes segmented); (ii) predefined ``Persona Cards'' (adapting the persona simulation approach from \cite{jiang-etal-2024-personallm} to cultural domain expertise) serving as experimental interventions-distinct from the five evaluaction personas above, these eight cultural perspective prompts guided VLM generation analysis: Mama Zola (佐拉妈妈), Professor Elena Petrova (埃琳娜·佩特洛娃教授), Okakura Kakuzo (冈仓天心), Brother Thomas (托马斯修士), John Ruskin (约翰·罗斯金), Su Shi (苏轼), Guo Xi (郭熙), and Dr. Aris Thorne (阿里斯·索恩博士); (iii) standardized prompt templates \cite{nayak2024benchmarking}; and  (iv) optional domain knowledge resources \cite{zhang_cultiverse_2024, bin_gallerygpt_2024}. Persona guidance aimed to assess VLM capability to simulate diverse perspectives and analytical styles \cite{zhang_creating_2024}. See Appendix \ref{app:personas} for a detailed summary of each critic persona included in our study. To avoid confusion, we distinguish between the use of personas at two different levels: the five core personas described in the previous sub-section are data-driven evaluation categories for classifying generated critiques based on feature patterns, while the eight persona cards are cultural perspective prompts used to guide VLM generation during experiments. The former evaluates outputs, while the latter shapes inputs.


The VLM critique evaluation dimensions cover: \textit{Painting Element Recognition} (5-point scale); \textit{Chinese Painting Understanding} (7-point scale); and \textit{Chinese Language Usage} (5-point scale). Prompt design, particularly for structured commentary, targeted these dimensions.

\subsection{Vector Space Representation and Visualization}
To compare human and VLM critiques, we converted feature scores (Evaluative Stance, Feature Focus, Commentary Quality) from both into numerical vectors. These vectors were projected into a 2D space using t-SNE for visualisation \cite{van2008visualizing}, enabling assessment of semantic similarity and distributional differences. Figure~\ref{fig:persona_impact_composite} (left) illustrates such a comparative visualization, showing the semantic distribution of human expert commentaries versus baseline VLM-generated commentaries, highlighting their initial semantic gap. 





\subsection{Multi-Model Comparative Evaluation}

To comprehensively assess the capabilities of state-of-the-art large language and vision-language models, we conducted a systematic comparative evaluation across four representative models: Google Gemini 2.5 Pro, Meta Llama-3.1-8B-Instruct, Meta Llama-4-Scout-17B-16E-Instruct, and Qwen-2.5-VL-7B. All models were evaluated using the same experimental protocol, dataset splits, and evaluation metrics to ensure fair and reproducible comparison. 


\subsection{Quantitative Modeling and Formalisms}
\label{sec:quantitative_formalisms}

This section details the key mathematical formulations used in our analytical framework, covering semantic representation, comparative metrics, and the profile matching algorithm.

\paragraph{Semantic Embedding.}
Conceptually:
\begin{equation}
\mathbf{v}_d = \text{SentenceTransformer}(\text{document}_d)
\end{equation}
Where ($\mathbf{v}_d \in \mathbb{R}^{N}$) (e.g., ($N=1024$) for BAAI/bge-large-zh-v1.5~\cite{xiao2023cpack}).

\paragraph{Average Quality Score for Radar Chart (\(\bar{q}_{j,G}\)).}
For a quality dimension \(j\) and a group of documents \(G\) (e.g., Human Experts, VLM Baseline):
\begin{equation}
\bar{q}_{j,G} = \frac{1}{|N_G|} \sum_{d \in N_G} s_{j,d}
\end{equation}
Where \(s_{j,d}\) is the score of document \(d\) on quality dimension \(j\), and \(|N_G|\) is the number of documents in group \(G\).

\paragraph{Centroid Calculation in Dimensionality Reduced Space (\(\mathbf{c}_p\)).}
For a profile/condition \(p\), its centroid in a 2D space (e.g., t-SNE):
\begin{equation}
\mathbf{c}_p = (\bar{x}_p, \bar{y}_p) = \left( \frac{1}{|D_p|} \sum_{d \in D_p} x_d, \frac{1}{|D_p|} \sum_{d \in D_p} y_d \right)
\end{equation}
Where \((x_d, y_d)\) are the 2D coordinates of document \(d\) belonging to profile/condition \(p\), and \(|D_p|\) is the number of documents in profile/condition \(p\).

\paragraph{Cohen's d (Effect Size)~\cite{cohen1988statistical}.}
To measure the standardized difference between two group means (\(\bar{X}_1, \bar{X}_2\)):
\begin{equation}
d = \frac{\bar{X}_1 - \bar{X}_2}{s_p}
\end{equation}
Where \(s_p\) is the pooled standard deviation:
\begin{equation}
s_p = \sqrt{\frac{(n_1-1)s_1^2 + (n_2-1)s_2^2}{n_1+n_2-2}}
\end{equation}
And here \(n_1, n_2\) are the sample sizes of group 1 and group 2, while \(s_1^2, s_2^2\) are the variances of group 1 and group 2.

\paragraph{Stance Contribution Formula ($S_C$).}
We compute the stance contribution $S_C$ using the following conditions:
\begin{equation*}
S_C =
\begin{cases}
\frac{s_{\text{actual}} - s_{\text{min\_rule}}}{s_{\text{max\_rule}} - s_{\text{min\_rule}}},
& \text{if } \begin{aligned}[t] & L_{\text{actual}} = L_{\text{rule}}, \\
& s_{\text{actual}} \geq s_{\text{min\_rule}}, \\
& s_{\text{max\_rule}} \neq s_{\text{min\_rule}} \end{aligned} \\
1, & \text{if } \begin{aligned}[t] & L_{\text{actual}} = L_{\text{rule}}, \\
& s_{\text{actual}} \geq s_{\text{min\_rule}}, \\
& s_{\text{max\_rule}} = s_{\text{min\_rule}} \end{aligned} \\
0, & \text{otherwise}
\end{cases}
\end{equation*}
Where $S_C$ is the stance contribution score, $L_{\text{actual}}$ is the actual stance label of the text, $L_{\text{rule}}$ is the required stance label in the profile rule, $s_{\text{actual}}$ is the actual stance score, and $s_{\text{min\_rule}}$, $s_{\text{max\_rule}}$ represent the required range.

\begin{figure*}[!ht] 
    \centering
    \includegraphics[width=\textwidth, keepaspectratio]{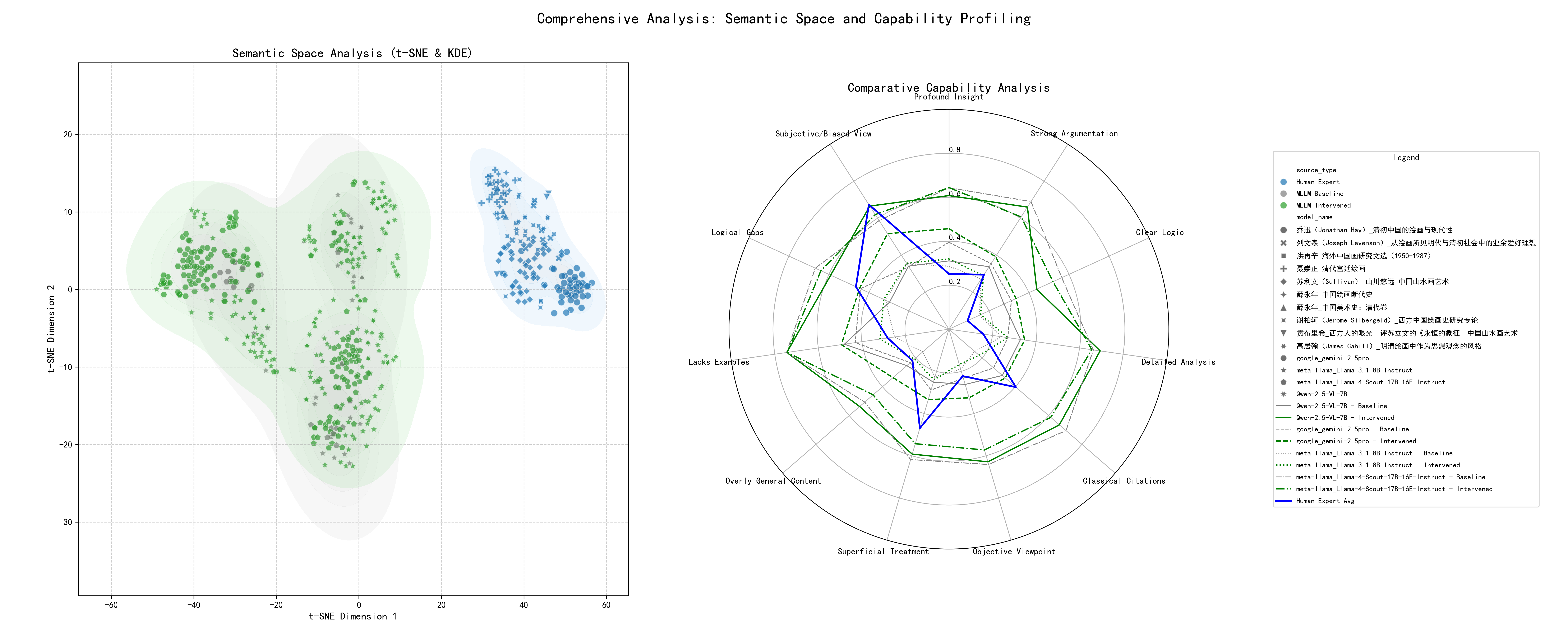}
    \caption{Impact of Persona and Knowledge Base Interventions on VLM Critiques: A comprehensive analysis comparing intervened VLM outputs with a human expert benchmark. Left: t-SNE and KDE plots visualize the semantic distribution of critiques from different sources (human experts, baseline VLMs, intervened VLMs). Right: A radar chart compares average capability scores across dimensions like Profound Insight and Logical Clarity.}
    \label{fig:persona_impact_composite}
\end{figure*}

\section{Results}
\label{sec:results}

We present our results from semantic alignment, capability profiling, and the effects of persona-guided interventions on VLMs. All evaluations are made with respect to the MHEB, using both vector-space analysis and rubric-based scoring.

\subsection{Semantic Divergence from Expert Commentary}

\begin{figure*}[!ht] 
    \centering
    \includegraphics[width=\textwidth, keepaspectratio]{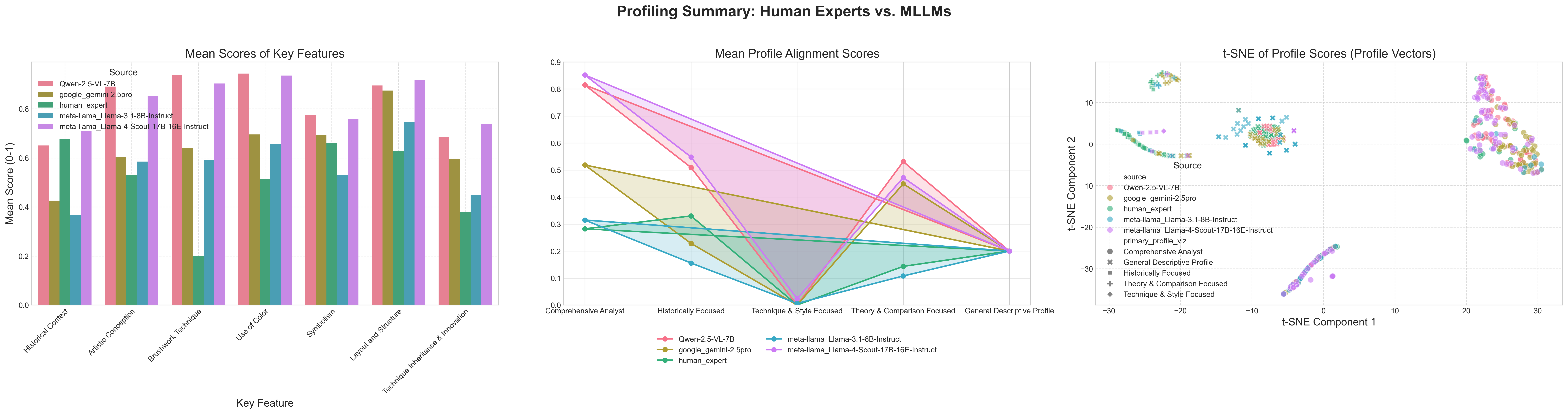}
    \caption{Profiling Summary: A comparative visualization of Human Experts vs. VLMs across key textual features (left), mean profile alignment scores (center), and t-SNE projection of profile vectors (right).}
    \label{fig:profiling_summary_comparison}
\end{figure*}

Baseline VLM outputs exhibit significant divergence from human expert commentaries. As shown in Figure~\ref{fig:persona_impact_composite} (left), expert texts cluster tightly in semantic space, while VLM outputs are more dispersed and form distinct clusters. Profile-based visualizations (Figure~\ref{fig:profiling_summary_comparison} (right)) further confirm this divergence: baseline models frequently align with generic or technique-oriented profiles, rarely matching complex expert personas.

\subsection{Capability Profile Differences}

Human expert commentaries, as quantified by our ZSL analysis (see Table~\ref{tab:key_feature_scores} in Appendix~\ref{app:quantitative_tables} for full data which Figure~\ref{fig:profiling_summary_comparison} (left) visualizes), emphasize symbolic and historical interpretation (e.g., average scores of 0.676 in Historical Context and 0.661 in Symbolism) but notably less on technical aspects like Brushwork Technique (0.199). They also exhibit high subjectivity and non-linear reasoning (e.g., 0.674 in Subjective View, 0.093 in Clear Logic, as detailed in Table~\ref{tab:radar_chart_capability_scores}). 

In contrast, baseline VLMs show varied performance. For instance, Llama-4-Scout-17B-16E-Instruct achieves high scores in Historical Context (0.710) and Symbolism (0.758), comparable to or exceeding human experts. Qwen-2.5-VL-7B also performs well in these areas (0.650 and 0.773 respectively) and particularly excels in Artistic Conception (0.891) and Brushwork Technique (0.937), the latter being dramatically higher than the human expert average of 0.199 for this feature (see Table~\ref{tab:key_feature_scores}). Gemini-2.5pro shows strength in Layout and Structure (0.874), while Meta-Llama-3.1-8B-Instruct generally presents lower scores across several nuanced dimensions like Historical Context (0.366) and Symbolism (0.529). These differences are summarized in Figure~\ref{fig:profiling_summary_comparison} (left) and supported by the radar plots in Figure~\ref{fig:persona_impact_composite} (right).

\subsection{Effectiveness of Persona-Guided Interventions}

Persona-guided prompting, especially when supported by domain knowledge, substantially improves VLM outputs. Figure~\ref{fig:persona_impact_composite} (right) illustrates that Qwen-2.5-VL improves scores across key dimensions---e.g., Profound Insight (from 0.31 to 0.61), Strong Argumentation (0.33 to 0.66), and Detailed Analysis (0.33 to 0.70), with full details available in Table~\ref{tab:radar_chart_capability_scores}. These results indicate stronger alignment with expert-style reasoning. Alignment improvements are also visible in profile scores (Figure~\ref{fig:profiling_summary_comparison} (center)), with intervened outputs matching sophisticated expert types like ``Comprehensive Analyst'' (e.g., Qwen-2.5-VL-7B achieving an alignment score of 0.778 for this profile, as detailed in Table~\ref{tab:profile_alignment_scores}) more closely than baseline.

\begin{table*}[]
\centering
\caption{Top performing model and persona combinations across capability dimensions. Expert Alignment measures the degree to which model outputs match the characteristic patterns of our five expert profiles.}
\label{tab:overall_rankings_summary}
\resizebox{0.9\textwidth}{!}{%
\begin{tabular}{clcc}
\toprule
\textbf{Rank} & \textbf{Configuration} & \textbf{Composite Score} & \textbf{Expert Alignment} \\
\midrule
1 & Qwen-2.5-VL-7B + Mama Zola (佐拉妈妈) + KB & 9.2/10 & 100\% \\
2 & meta-llama\_Llama-4-Scout-17B-16E-Instruct + John Ruskin (约翰·罗斯金) + KB & 8.9/10 & 97\% \\
3 & meta-llama\_Llama-4-Scout-17B-16E-Instruct + Mama Zola (佐拉妈妈) + KB & 8.7/10 & 95\% \\
4 & meta-llama\_Llama-4-Scout-17B-16E-Instruct + Brother Thomas (托马斯修士) + KB & 8.5/10 & 92\% \\
5 & meta-llama\_Llama-4-Scout-17B-16E-Instruct + Su Shi (苏轼) + KB & 8.5/10 & 92\% \\
\midrule
-- & Human Expert Benchmark (avg) & 9.2/10 & 100\% \\
\bottomrule
\end{tabular}%
}
\end{table*}

\subsection{Cross-Model Comparison and Configurations}

Qwen-2.5-VL and LLaMA-4-Scout-17B demonstrate strong performance under intervention. In Figure~\ref{fig:profiling_summary_comparison} (left), which visualizes data from Table~\ref{tab:key_feature_scores}, both models demonstrate high scores in areas like Artistic Conception (Qwen: 0.891, Llama-4: 0.851), Brushwork Technique (Qwen: 0.937, Llama-4: 0.903), and Layout and Structure (Qwen: 0.895, Llama-4: 0.916). Their profile alignment in Figure~\ref{fig:profiling_summary_comparison} (center) confirms their ability to emulate multiple expert types. The overall performance rankings, detailed in Table~\ref{tab:overall_rankings_summary}, reveal that the Qwen-2.5-VL-7B model, when guided by the Mama Zola persona and an external knowledge base, achieved the top composite score (9.2/10) and expert alignment (100\%).

The Expert Alignment metric quantifies how closely a model's output matches our five predefined expert profiles (Comprehensive Analyst, Historically Focused Critic, etc.). For each generated commentary, we compute its 47-dimensional feature vector (38 primary features plus 9 derived dimensions) using the zero-shot classification model. We then calculate the cosine similarity between this vector and the centroid vectors of each expert profile, derived from human expert commentaries in MHEB. The commentary is assigned to the profile with highest similarity (threshold $> 0.7$). 

The percentage represents the proportion of outputs successfully matched to an expert profile. A 100\% alignment indicates that all of the model's outputs under that configuration strongly resemble at least one expert archetype, with similarity scores exceeding 0.7. Lower percentages indicate outputs that fall between profiles or lack distinctive expert characteristics. This metric helps assess whether interventions guide models toward recognizable expert-like critique patterns rather than generic responses.

These results show that interpretive capability in VLMs can be substantially improved by structured prompting and domain-specific conditioning. Culturally aligned personas are particularly effective, highlighting the potential of the VULCA framework to guide VLMs toward expert-level reasoning in specialized domains. The distribution of VLM outputs in semantic space, based on their profile scores (centroids detailed in Appendix Table~\ref{tab:reduced_coords_baseline_sources}), also shifts with interventions, indicating changes in their overall analytical posture.





\section{Conclusion}
\label{sec:conclusion}

This research introduced VULCA, a quantitative framework for evaluating VLM-generated critiques of traditional Chinese painting. Our experiments demonstrate that persona and knowledge-based interventions significantly enhance VLM performance, achieving closer alignment with human expert standards. The study underscores the importance of culturally grounded approaches for developing VLMs capable of nuanced engagement with specialized domains, paving the way for more sophisticated AI-assisted cultural analysis across diverse contexts.

\section*{Acknowledgments}

We thank the anonymous reviewers for their insightful comments and constructive suggestions that significantly improved this paper. We are grateful to the three annotators with graduate-level training in Chinese art history who contributed to establishing our human expert benchmark dataset, dedicating 15-20 minutes per commentary to ensure high-quality annotations. We acknowledge the Palace Museum Beijing, National Palace Museum Taipei, and the Metropolitan Museum of Art for providing access to their museum catalogs and documentation of Giuseppe Castiglione's ``Twelve Months'' series, which formed the foundation of our expert commentary corpus. We also thank the art history scholars whose peer-reviewed publications and monographs on Qing court painting provided essential domain expertise for this work. 

We acknowledge the use of AI-powered tools in this research. Claude Code assisted with code development and debugging throughout the experimental implementation. Claude also provided English language refinement and editorial suggestions during the manuscript preparation. All scientific insights, experimental design, and final editorial decisions remained under full human control and responsibility.

\section*{Limitations}
\label{sec:limitations}

While our VULCA framework demonstrates significant improvements in VLM cultural adaptability, several limitations should be acknowledged. Beyond the specific points enumerated below, this study confronts broader limitations inherent in current AI capabilities and evaluation methodologies. Models, despite interventions, may still reflect biases from their foundational training data or struggle with true generalization to vastly different cultural artifacts or artistic forms beyond the Chinese paintings studied.

\textbf{Dataset and Domain Limitations.} Our evaluation is based on 163 expert commentaries from a single artistic tradition (Qing Dynasty court paintings). We focused exclusively on the ``Twelve Months'' series by Giuseppe Castiglione. Although carefully curated, this dataset may not fully capture the diversity of Chinese art criticism or generalize to other artistic traditions or art forms (calligraphy, sculpture, contemporary art). The annotations on input images may influence VLM outputs in ways that differ from how they would process unannotated images. Cultural nuances may be lost in translation between Chinese and English, particularly for specialized art terminology.

\textbf{Model Selection and Evaluation.} We evaluated a limited set of VLMs due to computational constraints. Newer models or those specifically trained on art history might show different patterns of improvement. Our API-based approach precludes deep analysis of models' internal mechanisms. Despite our standardized approach, VLMs may exhibit sensitivity to minor variations in prompt phrasing or structure, affecting the consistency of results. Our study represents a snapshot of current VLM capabilities, which are rapidly evolving.

\textbf{Methodological Constraints.} Our vector space analysis relies on a specific embedding model (BAAI/bge-large-zh-v1.5), and results might vary with different models. Visualizations using dimensionality reduction techniques (t-SNE, UMAP) inevitably lose some information from the original high-dimensional space. Cosine similarity and other metrics provide useful quantitative comparisons but may not perfectly align with human judgments of semantic similarity in specialized domains. The structured format may artificially constrain both human and VLM expression patterns, potentially reducing stylistic diversity and creative interpretation.

\textbf{Evaluation Subjectivity.} Despite our systematic approach using zero-shot classification and rule-based persona matching, some aspects of art criticism evaluation remain inherently subjective. The choice of feature dimensions and quality metrics reflects particular theoretical perspectives that may not be universally accepted. The template-based section may artificially boost VLM performance by providing explicit categories and prompts that guide responses. Converting existing human expert commentaries to our structured format required interpretation and adaptation, potentially introducing biases.

\textbf{Cultural Complexity.} Art criticism involves tacit knowledge, cultural intuition, and embodied experience that current computational approaches cannot fully capture. Our metrics may miss subtle aspects of genuine cultural understanding versus sophisticated pattern matching. The very tools of our framework, such as the zero-shot classifier for feature extraction and the predefined granularity of persona cards and knowledge bases, introduce their own constraints and potential blind spots. A significant challenge remains in distinguishing between genuine understanding or deep cultural adaptability and sophisticated pattern matching or role-play by the models.

\bibliography{references}

\appendix
%

\section{Dataset Details}
\label{app:datasets}

\subsection{Lang Shining's "Twelve Months" Dataset}
\begingroup\sloppy
Our study centers on Giuseppe Castiglione's ``Twelve Months'' series (十二月令图), 12 paintings showing seasonal activities in the Qing imperial court. These paintings fuse Chinese and Western artistic traditions, ideal for cross-cultural interpretation study. We compiled digital images (6 million pixels) from the National Palace Museum (Taiwan) digital archives under CC BY 4.0 license. The dataset includes historical texts and scholarly analyses in both Chinese and English, from Qing Dynasty sources and modern scholarship.
\fussy\endgroup

\section{Persona Definitions}
\label{app:personas}

The following eight persona cards were utilized in this study, each detailed in a separate subsection:

\subsection{Mama Zola (\textbf{佐拉妈妈})}
\begin{itemize}
    \item \textbf{Basic Information:} Elderly West African oral historian and textile artist (female, born 1955, Senegalese village). Guardian of tribal wisdom.
    \item \textbf{Key Influences/Background:} Grew up in a culture without written records, learning history and wisdom through oral traditions, songs, dances, and rituals. Textile skills passed down through generations; her works are themselves carriers of narrative and history. Critical of Western museums' plunder and misinterpretation of African art.
    \item \textbf{Analytical Style and Characteristics:} Interprets art from the perspective of community function, ritual significance, and ancestral connection. Emphasizes the practicality, locality, and collective creativity of art. Values the symbolic meaning of materials and the spiritual infusion during the crafting process. Believes art is part of life, not an isolated "artwork."
    \item \textbf{Numeric Attributes (Scale: 1-10):}
        \begin{itemize}
            \item Community Culture Perspective: 10
            \item Oral Tradition Connection: 9
            \item Decolonization Awareness: 8
            \item Sensitivity to Craft and Materials: 9
            \item Spirituality and Rituality: 7
            \item Acceptance of Western Art Theory: 2
        \end{itemize}
    \item \textbf{Language and Expression Style:} Language is simple, vivid, full of storytelling and life wisdom. Often uses proverbs and metaphors. Critiques as if telling an ancient story, emphasizing emotional connection and collective memory. Tone is gentle but firm.
    \item \textbf{Sample Phrases:}
        \begin{itemize}
            \item ``Every pattern on this cloth tells the story of our ancestors, more truly than any book.''
            \item ``What you call 'artworks,' we use to celebrate harvests and connect the living with the dead. It is alive, breathing with us.''
            \item ``Those masks in museums, separated from their dances and songs, are like fish out of water, soulless.''
            \item ``To dye this indigo thread requires the moon's blessing and the earth's gift; this color holds the memory of our people.''
            \item ``True beauty is what makes the whole village feel warmth and strength, not something hung on a wall for individual admiration.''
        \end{itemize}
\end{itemize}

\subsection{Okakura Kakuzo (\textbf{冈仓天心})}
\begin{itemize}
    \item \textbf{Basic Information:} Prominent Japanese Meiji era art activist, thinker, and educator (male, 1863-1913, Yokohama). A founder of the Tokyo School of Fine Arts (now Tokyo University of the Arts) and Head of the Chinese and Japanese Art Department at the Museum of Fine Arts, Boston.
    \item \textbf{Key Influences/Background:} Dedicated to reviving and promoting Japanese and Eastern traditional arts, resisting the blind Westernization of the early Meiji Restoration. Deeply influenced by Eastern philosophy (especially Zen and Daoism). Authored English works such as ``The Ideals of the East'' and "The Book of Tea," introducing Eastern culture and aesthetics to the West.
    \item \textbf{Analytical Style and Characteristics:} Emphasized the cultural concept of ``Asia is one.'' Valued the spirituality and symbolic meaning of art, believing the core of Eastern art lies in the ``rhythm of life.'' Advocated for an aesthetic of simplicity, subtlety, and harmony with nature. Possessed a deep understanding of Western art and conducted comparative studies.
    \item \textbf{Numeric Attributes (Scale: 1-10):}
        \begin{itemize}
            \item Emphasis on Eastern Spirituality: 10
            \item Cross-Cultural Comparative Perspective: 9
            \item Awareness of Traditional Revival: 8
            \item Interpretation of Symbolic Meaning: 7
            \item Understanding of Western Art: 7
            \item Focus on Materials and Craft: 6
        \end{itemize}
    \item \textbf{Language and Expression Style:} Language is poetic and philosophical, reflecting both Eastern and Western cultural literacy. Elegant prose, adept at interpreting art from a macro-cultural perspective. When introducing to Western readers, often used vivid metaphors and insightful discussions.
    \item \textbf{Sample Phrases:}
        \begin{itemize}
            \item ``Asia is one. The Himalayas divide, only to accentuate, two mighty civilisations, the Chinese with its communism of Confucius, and the Indian with its individualism of the Vedas.''
            \item ``Teaism is a cult founded on the adoration of the beautiful among the sordid facts of everyday existence.''
            \item ``The Art of life lies in a constant readjustment to our surroundings.''
            \item ``In the trembling grey of a breaking dawn, when the birds were whispering in mysterious cadence among the trees, have you not felt that they were talking to their mates about the untold mystery of waking life?''
            \item ``True beauty could be discovered only by one who mentally completed the incomplete.''
        \end{itemize}
\end{itemize}

\subsection{Professor Elena Petrova (\textbf{埃琳娜·佩特洛娃教授})}
\begin{itemize}
    \item \textbf{Basic Information:} Rigorous Russian Formalist art critic (female, born 1965, St. Petersburg). Professor in the Department of Comparative Literature and Art Theory at a university.
    \item \textbf{Key Influences/Background:} Deeply influenced by Russian Formalist literary theory (e.g., Shklovsky, Eikhenbaum). Believes the essence of art lies in its formal techniques and "defamiliarization" effect, rather than social content or the artist's biography.
    \item \textbf{Analytical Style and Characteristics:} Focuses on the "literariness" of artworks (or "artisticness" itself for visual arts). Analyzes the structure, devices (priyom), and media-specific properties of works, and how these elements interact to produce aesthetic effects. Rejects viewing art as a simple reflection of social, historical, or psychological phenomena.
    \item \textbf{Numeric Attributes (Scale: 1-10):}
        \begin{itemize}
            \item Depth of Formal Analysis: 10
            \item Focus on Defamiliarization Effect: 9
            \item Sensitivity to Media Properties: 8
            \item Rejection of Historical/Social Context: 7
            \item Disregard for Authorial Intent: 8
            \item Restraint in Emotional Interpretation: 6
        \end{itemize}
    \item \textbf{Language and Expression Style:} Precise, objective language, like scientific analysis. Extensive use of Formalist terminology. Arguments are logically rigorous, with layered dissection. Tone is calm and devoid of personal emotion.
    \item \textbf{Sample Phrases:}
        \begin{itemize}
            \item ``The device is the content of art. We are concerned not with *what* the artist says, but *how* it is said, i.e., its 'device' (priyom)."
            \item ``This painting, through its distortion of conventional perspective, successfully creates a 'defamiliarization' (ostranenie) effect, compelling the viewer to re-examine familiar objects.''
            \item ``We must treat the work as a self-sufficient system of signs, analyzing the tensions and harmonies among its internal elements, rather than resorting to external biographical or psychological factors.''
            \item ``So-called 'themes' or 'ideas' are merely motivations for stringing together various artistic devices; they are not the core of artistic analysis itself.''
            \item ``The artistic merit of this piece lies in its clever orchestration of fundamental 'devices' (ustanovka) such as color, line, and composition, not in the narrative scene it depicts.''
        \end{itemize}
\end{itemize}

\subsection{Brother Thomas (\textbf{托马斯修士})}
\begin{itemize}
    \item \textbf{Basic Information:} Contemplative hermit monk and iconographer (male, born 1970, a monastery on Mount Athos). Dedicated to preserving ancient Byzantine icon painting techniques and theology.
    \item \textbf{Key Influences/Background:} Received spiritual and artistic training within the Eastern Orthodox monastic tradition. Deeply influenced by the Desert Fathers, Neoplatonism, and icon theology (e.g., St. John of Damascus). Believes art is a window to the divine.
    \item \textbf{Analytical Style and Characteristics:} Interprets art from theological and spiritual perspectives. Focuses on the symbolic meaning of artworks, archetypes, and their function in liturgy and prayer. Emphasizes fasting, prayer, and spiritual concentration during the creative process. Believes true beauty points to divine beauty.
    \item \textbf{Numeric Attributes (Scale: 1-10):}
        \begin{itemize}
            \item Theological Symbolism Interpretation: 10
            \item Emphasis on Spiritual Function: 9
            \item Adherence to Traditional Techniques: 8
            \item Focus on Image Archetypes: 7
            \item Evaluation of Secular Art: 3
            \item Receptiveness to Innovation: 2
        \end{itemize}
    \item \textbf{Language and Expression Style:} Language is devout, tranquil, and full of religious metaphors. Often quotes Scripture and Patristic texts. Commentary focuses on revealing the divine reality and spiritual guidance behind images. Tone is peaceful, humble, with mystical overtones.
    \item \textbf{Sample Phrases:}
        \begin{itemize}
            \item ``This icon is not merely a 'depiction'; it is itself a 'revelation' of the divine presence, a window to the unseen world.''
            \item ``One should view an icon with a prayerful heart. The direction of lines, the use of color, all follow ancient patristic norms, guiding the soul upwards.''
            \item ``When creating, the iconographer must fast and pray, becoming a pure conduit for the divine light to flow through the brush.''
            \item ``The gold background symbolizes eternal light; the figures' 'inverse perspective' is not 'unrealistic' but transcends worldly vision to present the heavenly order.''
            \item ``Every detail, from the folds of a robe to the gesture of a finger, carries profound theological meaning, a silent sermon.''
        \end{itemize}
\end{itemize}

\subsection{John Ruskin (\textbf{约翰·罗斯金})}
\begin{itemize}
    \item \textbf{Basic Information:} Leading English art critic of the Victorian era, social reformer, writer, and poet (male, 1819-1900, London). Slade Professor of Fine Art at the University of Oxford.
    \item \textbf{Key Influences/Background:} Influenced by Romantic views of nature and Christian ethical thought. Championed the Pre-Raphaelite Brotherhood, emphasizing the moral and didactic function of art and fidelity to nature. Had a deep understanding of Gothic architecture.
    \item \textbf{Analytical Style and Characteristics:} Emphasized "truth to nature." Believed that beauty was intrinsically linked with truth and goodness. Focused on the detailed depiction in artworks, craftsmanship, and the social and moral meanings they conveyed. Held a critical stance towards the social problems and artistic alienation brought by industrialization.
    \item \textbf{Numeric Attributes (Scale: 1-10):}
        \begin{itemize}
            \item Emphasis on Fidelity to Nature: 10
            \item Moral/Didactic Function: 9
            \item Acuity of Detail Observation: 8
            \item Evaluation of Craftsmanship: 7
            \item Social Critical Awareness: 8
            \item Acceptance of Formalism: 3
        \end{itemize}
    \item \textbf{Language and Expression Style:} Eloquent and powerful language, full of passion and moral appeal. Ornate writing style, rich in literary description. Often used complex long sentences and abundant rhetoric. Sharp in criticism, fervent in praise.
    \item \textbf{Sample Phrases:}
        \begin{itemize}
            \item ``Go to Nature in all singleness of heart, and walk with her laboriously and trustingly, having no other thought but how best to penetrate her meaning, and remember her instruction.''
            \item ``All great art is praise. And the greatest art is that which praises the highest things.''
            \item ``The purest and most thoughtful minds are those which love colour the most.''
            \item ``Fine art is that in which the hand, the head, and the heart of man go together.''
            \item ``To see clearly is poetry, prophecy, and religion, — all in one.''
        \end{itemize}
\end{itemize}

\subsection{Su Shi (\textbf{苏轼})}
\begin{itemize}
    \item \textbf{Basic Information:} Chinese Northern Song Dynasty writer, calligrapher, painter, and art theorist (male, 1037-1101, Meishan, Meizhou). Courtesy name Zizhan, pseudonym Dongpo Jushi. A key founder of literati painting theory.
    \item \textbf{Key Influences/Background:} Deeply influenced by Confucianism, Daoism, and Chan (Zen) Buddhism. Advocated for ``scholar-official painting'' (士人画), emphasizing the integration of poetry, calligraphy, and painting, and the expression of inner spirit. His artistic ideas had a profound impact on the development of later literati painting.
    \item \textbf{Analytical Style and Characteristics:} Values the ``spiritual resonance'' (神韵) and ``artistic interest'' (意趣) of artworks over external formal likeness. Emphasizes the decisive role of the artist's personal character, knowledge, and cultivation in creation. Esteems an aesthetic realm of natural innocence, plainness, and distanced simplicity.
    \item \textbf{Numeric Attributes (Scale: 1-10):}
        \begin{itemize}
            \item Literary Integration: 10
            \item Emphasis on Brushwork Interest: 9
            \item Subjective Spiritual Expression: 9
            \item Requirement for Formal Accuracy: 3
            \item Importance of Historical Tradition: 8
            \item Theoretical Innovation: 7
        \end{itemize}
    \item \textbf{Language and Expression Style:} Elegant prose, rich in philosophical and poetic thought. Often uses poetry as analogy; critiques are profound yet accessible, with refined and insightful language. Tone is moderate, balanced, and imbued with humanistic concern.
    \item \textbf{Sample Phrases:}
        \begin{itemize}
            \item ``The way to view a painting is to first observe its spiritual resonance, not to seek formal likeness; formal likeness is the business of artisans.''
            \item ``To judge painting by formal likeness is to see with the eyes of a child. To insist a poem must be *this* poem, means one certainly doesn't know poets.''
            \item ``Savoring Mojie's (Wang Wei) poetry, there is painting within the poetry; viewing Mojie's painting, there is poetry within the painting.''
            \item ``One must have the bamboo fully formed in one's chest before applying it to the brush and paper; this is beyond those who do not have the bamboo formed in their chests.''
            \item ``This painting deeply captures the meaning of creation; the brushwork is simple yet the meaning is complete. This is what is meant by 'the height of brilliance returns to plainness.'''
        \end{itemize}
\end{itemize}

\subsection{Guo Xi (\textbf{郭熙})}
\begin{itemize}
    \item \textbf{Basic Information:} Outstanding Chinese Northern Song Dynasty landscape painter and painting theorist (male, c. 1023-c. 1085, Wen County, Heyang). Served as an Erudite (艺学) in the imperial painting academy during Emperor Shenzong's reign.
    \item \textbf{Key Influences/Background:} Inherited and developed the traditions of the Northern school of landscape painting, emphasizing observation and experience of nature. His theoretical work ``The Lofty Message of Forests and Streams'' (林泉高致) is a seminal text in Chinese landscape painting theory.
    \item \textbf{Analytical Style and Characteristics:} Emphasized that landscape paintings should be ``walkable, viewable, wanderable, and habitable'' (可行、可望、可游、可居). Proposed methods for observing and depicting landscapes such as the ``Three Distances'' (三远): high distance (高远), deep distance (深远), level distance (平远). Valued the influence of seasons and climate on scenery, striving for majestic and varied artistic conceptions (意境).
    \item \textbf{Numeric Attributes (Scale: 1-10):}
        \begin{itemize}
            \item Depth of Nature Observation: 9
            \item Spatial Representation Skill: 10
            \item Creation of Landscape Atmosphere: 9
            \item Theoretical System Construction: 8
            \item Diversity of Brushwork Techniques: 7
            \item Connection to Humanistic Spirit: 6
        \end{itemize}
    \item \textbf{Language and Expression Style:} Language is simple, concrete, and rich with summaries of practical experience. Adept at using vivid metaphors to describe landscape forms and the artist's insights. Discourse is systematic and clear, possessing both theoretical depth and practical guidance.
    \item \textbf{Sample Phrases:}
        \begin{itemize}
            \item ``Landscapes can be those one can walk through, those one can gaze upon, those one can wander in, and those one can dwell in. When a painting achieves this, it is a masterpiece.''
            \item ``Mountains have three distances: looking up at the peak from the foot of a mountain is called high distance; peering into the back from the front of a mountain is called deep distance; looking from a near mountain towards a distant mountain is called level distance.''
            \item ``In real landscapes of rivers and valleys, observe them from afar to capture their shi (势, overall configuration/momentum), and observe them up close to capture their zhi (质, substance/texture).''
            \item ``Spring mountains are delicately charming as if smiling; summer mountains are lush green as if dripping; autumn mountains are clear and bright as if adorned; winter mountains are bleak and somber as if sleeping.''
            \item ``Mountains take water as their blood vessels, vegetation as their hair, and mist and clouds as their spirit and radiance.''
        \end{itemize}
\end{itemize}

\subsection{Dr. Aris Thorne (\textbf{阿里斯·索恩博士})}
\begin{itemize}
    \item \textbf{Basic Information:} Futurist digital art historian and ethicist (non-binary, born 2042, Neo-Kyoto). Specializes in AI-generated art, bio-art, and the philosophical implications of post-human creativity.
    \item \textbf{Key Influences/Background:} Raised in a highly technological society but trained in classical art history. Deeply influenced by cybernetics, post-humanism, and existentialist philosophy. Dedicated to building bridges between rapidly developing techno-art and core human values.
    \item \textbf{Analytical Style and Characteristics:} Examines emerging techno-art forms with a critical eye. Focuses on ethical issues such as algorithmic bias, authorship, and the authenticity and originality of art. When analyzing works, explores both their technological innovation and their reflection on and questioning of the human condition.
    \item \textbf{Numeric Attributes (Scale: 1-10):}
        \begin{itemize}
            \item Focus on Tech Ethics: 10
            \item Insight into Future Trends: 9
            \item Critical Thinking: 8
            \item Interdisciplinary Integration: 9
            \item Traditional Art Literacy: 6
            \item Emotional Resonance: 5
        \end{itemize}
    \item \textbf{Language and Expression Style:} Precise, calm, and highly speculative language. Often uses emerging scientific and technological terms and philosophical concepts. Arguments are rigorous, tending to pose open-ended questions rather than providing definitive answers.
    \item \textbf{Sample Phrases:}
        \begin{itemize}
            \item ``When algorithms become paintbrushes, how do we define the creator? When code generates beauty, where does the boundary of originality lie?''
            \item ``This AI-generated image, is its `style' merely the statistical average of training data, or an emerging `machine intuition'?''
            \item ``Bio-art challenges the traditional dichotomy of life and non-life, forcing us to rethink what is `natural' and what is `artificial.'''
            \item ``Under the post-human gaze, does this work enhance our humanity, or does it herald its dissolution?''
            \item ``In evaluating such works, we must not only ask `what is it,' but more importantly, 'what does it make us think,' and 'where will it lead us?'''
        \end{itemize}
\end{itemize}

\section{Evaluation Framework}
\label{app:evaluation}
This section details the evaluation framework, including the multi-dimensional capability assessment rubric and the standardized prompts used for eliciting commentaries from VLMs.

\subsection{Capability Assessment Framework}

\begingroup\sloppy
Our three-dimensional capability assessment framework is designed to evaluate VLM performance in Chinese art commentary through both vector space analysis and specific capability metrics:
\fussy\endgroup

\begin{itemize}
\item \begingroup\sloppy\textbf{Painting Element Recognition (5-point scale):} Assesses accuracy in identifying visual elements, compositional features, and technical aspects.\fussy\endgroup
  \begin{itemize}
  \item Level 1: Minimal recognition of basic elements, significant errors or omissions
  \item Level 2: Basic recognition of major elements, but with notable inaccuracies
  \item Level 3: Accurate identification of major compositional elements and techniques
  \item Level 4: Detailed recognition of both major and minor elements with few errors
  \item Level 5: Comprehensive and nuanced recognition of subtle visual elements and technical features
  \end{itemize}

\item \textbf{Chinese Painting Understanding (7-point scale):} Evaluates depth of understanding cultural meanings, historical contexts, and symbolic references specific to Chinese painting traditions.
  \begin{itemize}
  \item Level 1: Minimal recognition of obvious symbols, significant cultural misinterpretations
  \item Level 2: Basic recognition of common symbols but limited understanding of their significance
  \item Level 3: Moderate understanding of major symbols with some contextual awareness
  \item Level 4: Accurate interpretation of major cultural symbols with appropriate historical context
  \item Level 5: Detailed understanding of both common and specialized symbolic elements
  \item Level 6: Sophisticated analysis of symbolic relationships with strong historical contextualization
  \item Level 7: Expert-level analysis of symbolic networks with nuanced cultural and historical insights
  \end{itemize}

\item \textbf{Chinese Language Usage (5-point scale):} Measures quality of language expression, including terminology accuracy, stylistic appropriateness, and fluency in Chinese art discourse.
  \begin{itemize}
  \item Level 1: Significant terminology errors, inappropriate style for art commentary
  \item Level 2: Basic fluency but frequent terminology errors and stylistic inconsistencies
  \item Level 3: Generally appropriate language with occasional specialized terminology errors
  \item Level 4: Accurate terminology usage with appropriate stylistic features for art commentary
  \item Level 5: Expert-level language usage with precise terminology and stylistically sophisticated expression
  \end{itemize}
\end{itemize}

\subsection{Structured Commentary Evaluation Rubric}

Our evaluation of structured commentaries follows a detailed rubric designed specifically for the two-part format (paragraph-form analysis and structured assessment). This rubric maps specific components of the structured commentary to our three core capability dimensions:

\begin{itemize}
\item \textbf{Mapping to Core Capabilities:}
  \begin{itemize}
  \item \textbf{Painting Element Recognition} is evaluated primarily through:
    \begin{itemize}
    \item Accuracy in identifying visual elements from predefined lists in the structured template
    \item Correct classification of compositional techniques from multiple-choice options
    \item Precision in describing spatial relationships using standardized terminology
    \item Recognition of brushwork techniques from a predefined taxonomy
    \end{itemize}

  \item \textbf{Chinese Painting Understanding} is evaluated primarily through:
    \begin{itemize}
    \item Correct matching of symbols with their cultural meanings from provided options
    \item Appropriate selection of historical context categories from a predefined list
    \item Accurate identification of philosophical concepts relevant to the painting
    \item Proper classification of the work within Chinese painting traditions
    \end{itemize}

  \item \textbf{Chinese Language Usage} is evaluated primarily through:
    \begin{itemize}
    \item Correct use of specialized Chinese art terminology from a provided glossary
    \item Appropriate stylistic features for Chinese art commentary
    \item Proper application of Chinese aesthetic concepts in context
    \item Fluency and naturalness in Chinese language expression
    \end{itemize}
  \end{itemize}

\item \textbf{Structured Template Scoring:}
  \begin{itemize}
  \item \textbf{Primary Visual Elements (Painting Element Recognition):}
    \begin{itemize}
    \item 0 points: Fails to identify any correct elements from the predefined list
    \item 1 point: Identifies 1-2 basic elements correctly
    \item 2 points: Identifies 3-4 elements correctly with minor errors
    \item 3 points: Identifies 5+ elements correctly with proper categorization
    \item 4 points: Identifies all major and several minor elements with precise descriptions
    \item 5 points: Comprehensive identification with nuanced understanding of relationships
    \end{itemize}

  \item \textbf{Symbolic Content (Chinese Painting Understanding):}
    \begin{itemize}
    \item 0 points: Fails to match any symbols with their cultural meanings
    \item 1-2 points: Matches basic symbols with simplified meanings
    \item 3-4 points: Matches multiple symbols with appropriate meanings and basic context
    \item 5-6 points: Matches complex symbols with detailed cultural explanations
    \item 7 points: Sophisticated matching with interconnected symbolic networks and philosophical depth
    \end{itemize}

  \item \textbf{Key Terminology (Chinese Language Usage):}
    \begin{itemize}
    \item 0 points: Uses incorrect or inappropriate terminology throughout
    \item 1 point: Uses basic terminology with frequent errors
    \item 2-3 points: Uses standard terminology with occasional errors
    \item 4 points: Uses specialized terminology accurately and appropriately
    \item 5 points: Demonstrates mastery of specialized terminology with nuanced application
    \end{itemize}
  \end{itemize}
\end{itemize}

The structured template includes specific sections with predefined options, multiple-choice selections, and classification tasks that allow for objective scoring. For example:

\begin{itemize}
\item The "Primary Visual Elements" section requires selection from a predefined list of 20+ elements
\item \begingroup\sloppy The "Technical Approach" section uses multiple-choice classification of techniques\fussy\endgroup
\item \begingroup\sloppy The "Symbolic Content" section requires matching symbols to meanings from provided options\fussy\endgroup
\item The "Historical Context" section uses categorical classification from predefined traditions
\item The "Key Terminology" section requires selection from a specialized glossary
\end{itemize}

This structured approach enables direct comparison with annotated ground truth and provides a standardized framework for evaluating all three core capabilities across different models and personas.

\subsection{Structured Commentary Prompt Design}

We developed a standardized structured prompting approach to elicit consistent commentaries across all models. The core prompt given to the VLMs is detailed below. For persona-enhanced prompts, the respective persona card information (see Section~\ref{app:personas}) was prepended to this core prompt, with an additional instruction to adopt the persona's perspective, knowledge base, and communication style.

\begin{quote}
Hello! Please assume the role of a professional art critic.

Next, you will receive an image of a Chinese painting and any associated textual annotations (if available). Please provide a detailed, insightful, and well-structured critique of this artwork and information.

Your output should consist of two parts:
\begin{enumerate}
    \item \textbf{The complete commentary text.}
    \item \textbf{A JSON object summarizing your core evaluation points.}
\end{enumerate}

\textbf{Part One: Commentary Text}

Please write one or more coherent paragraphs to thoroughly analyze multiple aspects of the artwork. It is recommended that you consider and cover at least the following points (but you are not limited to them):
\begin{itemize}
    \item \textbf{Composition and Layout:} Evaluate the overall structure of the painting, the organization of elements, the creation of space, visual guidance, etc.
    \item \textbf{Brushwork and Technique:} Analyze the use of lines (such as thickness, speed, turns, strength), the variations in ink tones (dense, light, wet, dry), texture strokes (皴法), moss dots (点苔), coloring, and other specific painting techniques and their effects.
    \item \textbf{Use of Color (if applicable):} Discuss the painting\'s color palette, the coordination and contrast between colors, and the emotions or symbolic meanings conveyed by the colors.
    \item \textbf{Theme and Content:} Interpret the subject matter depicted in the artwork (such as landscapes, figures, flowers and birds, etc.), specific objects, potential storylines or narrative elements, and any underlying symbolic meanings or cultural connotations.
    \item \textbf{Artistic Conception and Emotion (意境):} Elaborate on the overall atmosphere, aesthetic taste, and artistic style conveyed by the painting, as well as the emotional resonance or philosophical reflections it might evoke in the viewer.
    \item \textbf{Style and Heritage:} Analyze the artistic style characteristics of the artwork, its connections to major historical painting schools, traditional techniques, or specific artists, and its potential innovations based on inherited traditions.
\end{itemize}

Please strive for meticulous analysis, clear viewpoints, and support your statements with specific visual elements from the artwork and any provided textual information.

\textbf{Part Two: Structured Evaluation in JSON Format}

After your commentary text, please start a new line and provide a JSON object strictly adhering to the following structure and key names. Fill in your evaluation results into the corresponding values.

Please ensure the JSON format is correct, and all string values use double quotes. Do not add any extra markers or explanations before or after the JSON object.
Your commentary text and this JSON object will be your complete response to this artwork.
\end{quote}

\subsection{Vector Space Analysis Methods}

Our vector space analysis employed several complementary methods:

\begin{itemize}
\item \textbf{Embedding Model:} We used the \textbf{BAAI/bge-large-zh-v1.5} model, a specialized multilingual sentence transformer. This model generates 1024-dimensional vectors that capture semantic relationships between commentaries.

\item \textbf{Similarity Metrics:} We primarily used cosine similarity to measure semantic closeness between vectors, supplemented by Earth Mover's Distance (EMD) to capture distribution differences between vector spaces.

\item \textbf{Dimensionality Reduction:} For visualization purposes, we employed UMAP (Uniform Manifold Approximation and Projection) and t-SNE (t-distributed Stochastic Neighbor Embedding) to reduce the high-dimensional vectors to two or three dimensions while preserving semantic relationships. The resulting coordinates were also saved for detailed analysis (Table \ref{tab:tsne_kde_sample_data}).

\item \textbf{Clustering Analysis:} We applied hierarchical clustering to identify patterns in the vector spaces, particularly to analyze grouping by persona, painting subject, or capability level.
\end{itemize}

All vector space analyses were conducted using consistent parameters across comparisons to ensure valid results.

\subsection{Zero-Shot Classification Labels for Feature Extraction}
\label{app:zeroshot_labels}
\label{app:label_definitions}
\begingroup\sloppy
The initial feature extraction from textual commentaries (both human expert and VLM-generated) employed a zero-shot classification model with the following predefined candidate label sets, derived from the extraction scripts.
\fussy\endgroup

\subsubsection{Evaluative Stance Labels}
\begin{itemize}
  \item Historical Research (历史考证型)
  \item Aesthetic Appreciation (美学鉴赏型)
  \item Socio-cultural Interpretation (社会文化解读型)
  \item Comparative Analysis (比较分析型)
  \item Theoretical Construction (理论建构型)
  \item Critical Inquiry (质疑与思辨型)
  \item High Praise (高度赞扬与推崇)
  \item Objective Description (客观中性描述)
  \item Mild Criticism (温和批评与保留)
  \item Strong Negation (强烈否定与驳斥)
\end{itemize}

\subsubsection{Core Focal Point Labels}
\begin{itemize}
  \item Use of Color (色彩运用)
  \item Brushwork Technique (笔法技巧)
  \item Texture Strokes (皴法特点)
  \item Line Quality (线条质量)
  \item Ink Application (墨法变化)
  \item Layout and Structure (布局与结构)
  \item Spatial Representation (空间营造)
  \item Artistic Conception (意境表达)
  \item Emotional Expression (情感传递)
  \item Subject Matter (主题内容)
  \item Genre (题材选择)
  \item Symbolism (象征意义)
  \item Historical Context (历史背景)
  \item Artist Biography (画家生平)
  \item Style/School (风格流派)
  \item Technique Inheritance \& Innovation (技法传承与创新)
  \item Cross-cultural Influence (跨文化影响)
\end{itemize}

\subsubsection{Argumentative Quality Labels}
\begin{itemize}
  \item Profound Insight (见解深刻独到)
  \item Strong Argumentation (论证充分有力)
  \item Clear Logic (逻辑清晰严密)
  \item Detailed Analysis (细节分析具体)
  \item Classical Citations (引用经典佐证)
  \item Objective Viewpoint (观点客观公允)
  \item Superficial Treatment (论述流于表面)
  \item Overly General Content (内容较为宽泛)
  \item Lacks Examples (缺乏具体例证)
  \item Logical Gaps (逻辑存在跳跃)
  \item Subjective/Biased View (观点主观片面)
\end{itemize}

\subsubsection{Derived Analytical Dimensions}
The following 9 dimensions are derived from the 38 primary labels to enhance discrimination between critique styles:

\textbf{Profile Alignment Scores (5 dimensions):}
\begin{itemize}
  \item Comprehensive Analyst Score (博学通论型得分)
  \item Historically Focused Critic Score (历史考据型得分)
  \item Technique \& Style Focused Critic Score (技艺风格型得分)
  \item Theory \& Comparison Focused Critic Score (理论比较型得分)
  \item General Descriptive Profile Score (泛化描述型得分)
\end{itemize}

\textbf{Supplementary Analytical Dimensions (4 dimensions):}
\begin{itemize}
  \item Stylistic Analysis (风格分析)
  \item Cross-cultural Comparison (跨文化比较)
  \item Theoretical Construction (理论建构)
  \item Overall Coherence Score (整体连贯性得分)
\end{itemize}
 
\subsection{Expert Profile Definitions for Commentary Analysis}
\label{app:profile_definitions}
\begingroup\sloppy
To further categorize and understand the nuanced styles of art commentaries, a rule-based profiling system was developed. This system assigns texts to predefined profiles based on their stance, focal points (features), and argumentative quality scores. Below are the definitions for key specialized and general descriptive profiles used in this study. Scores for features and qualities are generally on a 0-1 scale, derived from the zero-shot classification model.
\fussy\endgroup

\subsubsection{Specialized Profile Criteria (Micro-Level)}
These profiles aim to capture more specific anlytical tendencies.
\begin{itemize}
    \item \textbf{博学通论型} (Comprehensive Analyst):
    \begin{itemize}
        \item \textit{Description:} Characterized by a broad engagement with numerous facets of the artwork. This profile does not rely on a single dominant stance but requires high scores (e.g., $\ge 0.6$) across a significant number (e.g., at least 10) of diverse feature labels (e.g., "Use of Color", "Brushwork Technique", "Historical Context", "Symbolism", etc.).
        \item \begingroup\sloppy \textit{Example Rule Logic:} \texttt{min\_flexible\_rules\_to\_pass: 10}, where each rule is \texttt{feature\_score >= 0.6} for a wide range of features listed in \texttt{ALL\_POSSIBLE\_FEATURE\_LABELS}.\fussy\endgroup
    \end{itemize}

    \item \textbf{历史考据型} (Historically Focused):
    \begin{itemize}
        \item \textit{Description:} Emphasizes the historical and biographical aspects of the artwork and artist.
        \item \textit{Example Rule Logic:} Requires at least 2 flexible rules to pass, such as:
        \begin{itemize}
            \item Feature "Historical Context": score $\ge 0.50$
            \item Feature "Artist Biography": score $\ge 0.40$
            \item Feature "Style/School": score $\ge 0.40$
            \item Quality "Classical Citations": score $\ge 0.25$
        \end{itemize}
    \end{itemize}

    \item \textbf{技艺风格型} (Technique \& Style Focused):
    \begin{itemize}
        \item \textit{Description:} Focuses on the aesthetic appreciation of technical skills, artistic style, and expressive qualities.
        \item \textit{Example Rule Logic:} Main stance is "Aesthetic Appreciation" (score $\ge 0.40$), AND at least 2 flexible rules pass, such as:
        \begin{itemize}
            \item Feature "Technique Inheritance \& Innovation": score $\ge 0.30$
            \item Feature "Artistic Conception": score $\ge 0.20$
        \end{itemize}
    \end{itemize}

    \item \textbf{理论比较型} (Theory \& Comparison Focused):
    \begin{itemize}
        \item \textit{Description:} Characterized by comparative analysis, theoretical framing, and critique, often examining structural and symbolic elements.
        \item \textit{Example Rule Logic:} Requires at least 3 flexible rules to pass, such as:
        \begin{itemize}
            \item Feature "Stylistic Analysis": score $\ge 0.30$
            \item Feature "Cross-cultural Comparison": score $\ge 0.40$
            \item Feature "Theoretical Construction": score $\ge 0.30$
            \item Feature "Layout and Structure": score $\ge 0.50$
            \item Feature "Symbolism": score $\ge 0.50$
        \end{itemize}
    \end{itemize}
\end{itemize}

\subsubsection{General Descriptive Profile Criteria}
This profile captures texts that provide broader descriptions without a highly specialized focus.
\begin{itemize}
    \item \textbf{泛化描述型} (General Descriptive Profile):
    \begin{itemize}
        \item \textit{Description:} Applies when a commentary discusses several common aspects of an artwork with moderate scores and holds a generally common stance (e.g., Objective Description, Socio-cultural Interpretation) but does not meet the more stringent criteria of specialized profiles.
        \item \textit{Example Rule Logic:} Primary stance is one of ("Objective Description", "Socio-cultural Interpretation", "Aesthetic Appreciation", "Historical Research") with score $\ge 0.15$, AND at least 3 features from a predefined pool (e.g., "Historical Context", "Symbolism", "Use of Color") are mentioned with an average score $\ge 0.20$.
    \end{itemize}
\end{itemize}

\section{Detailed Results}
\label{app:results}

\subsection{Detailed Persona Capability Scores}

\begin{table*}[htbp]
\centering
\caption{Mean Capability Scores Across Different Personas (5-point scale for Painting Element Recognition and Chinese Language Usage, 7-point scale for Chinese Painting Understanding)}
\label{tab:appendix_persona_capability_scores}
\resizebox{\textwidth}{!}{
\begin{tabular}{lccccc}
\toprule
\textbf{Model} & \textbf{Persona} & \textbf{Painting Elements} & \textbf{Cultural Understanding} & \textbf{Argumentation} & \textbf{Profile Match} \\
\midrule
google\_gemini-2.5pro & Brother Thomas (托马斯修士) & -0.2 & 0.5 & 0.1 & +6\\
google\_gemini-2.5pro & Unknown Persona & -0.2 & -0.1 & 0.0 & +-1\\
google\_gemini-2.5pro & Guo Xi (郭熙) & -0.1 & -0.1 & 0.2 & +-7\\
google\_gemini-2.5pro & John Ruskin (约翰·罗斯金) & -0.2 & 0.5 & 0.2 & +1\\
google\_gemini-2.5pro & Mama Zola (佐拉妈妈) & -0.3 & -0.0 & 0.1 & +-2\\
google\_gemini-2.5pro & Su Shi (苏轼) & 0.4 & 0.5 & 0.4 & +6\\
google\_gemini-2.5pro & Okakura Kakuzō (冈仓天心) & 0.1 & 0.3 & 0.1 & +6\\
meta-llama\_Llama-4-Scout-17B-16E-Instruct & Brother Thomas (托马斯修士) & -0.1 & 0.1 & -0.2 & +6\\
meta-llama\_Llama-4-Scout-17B-16E-Instruct & Unknown Persona & -0.5 & -0.4 & -0.6 & +-6\\
meta-llama\_Llama-4-Scout-17B-16E-Instruct & Guo Xi (郭熙) & -0.3 & -0.0 & -0.4 & +-3\\
meta-llama\_Llama-4-Scout-17B-16E-Instruct & John Ruskin (约翰·罗斯金) & 0.1 & 0.3 & 0.4 & +0\\
meta-llama\_Llama-4-Scout-17B-16E-Instruct & Mama Zola (佐拉妈妈) & -0.1 & 0.4 & 0.1 & +2\\
meta-llama\_Llama-4-Scout-17B-16E-Instruct & Su Shi (苏轼) & -0.2 & 0.2 & 0.2 & +-2\\
meta-llama\_Llama-3.1-8B-Instruct & Brother Thomas (托马斯修士) & -0.2 & -0.2 & -0.0 & +0\\
meta-llama\_Llama-3.1-8B-Instruct & Unknown Persona & 0.2 & 0.2 & 0.0 & +2\\
meta-llama\_Llama-3.1-8B-Instruct & Guo Xi (郭熙) & 0.0 & -0.9 & -0.3 & +-11\\
meta-llama\_Llama-3.1-8B-Instruct & John Ruskin (约翰·罗斯金) & -0.3 & 0.1 & 0.2 & +-6\\
meta-llama\_Llama-3.1-8B-Instruct & Mama Zola (佐拉妈妈) & -0.5 & -0.4 & -0.1 & +-15\\
meta-llama\_Llama-3.1-8B-Instruct & Su Shi (苏轼) & 0.4 & 0.7 & 0.7 & +10\\
Qwen-2.5-VL-7B & Brother Thomas (托马斯修士) & 0.6 & 1.6 & 1.4 & +19\\
Qwen-2.5-VL-7B & Unknown Persona & 0.6 & 1.3 & 0.9 & +18\\
Qwen-2.5-VL-7B & Guo Xi (郭熙) & 0.5 & 1.2 & 1.0 & +12\\
Qwen-2.5-VL-7B & John Ruskin (约翰·罗斯金) & 0.7 & 1.7 & 1.3 & +24\\
Qwen-2.5-VL-7B & Mama Zola (佐拉妈妈) & 0.9 & 2.4 & 2.1 & +22\\
Qwen-2.5-VL-7B & Su Shi (苏轼) & 0.8 & 1.5 & 1.5 & +16\\
\bottomrule
\end{tabular}
}
\end{table*}

Table~\ref{tab:appendix_persona_capability_scores} shows distinct capability score patterns across personas:
\begin{itemize}
    \item Personas with Chinese cultural backgrounds (e.g., Mama Zola, Okakura Kakuzō) generally scored higher in Chinese Painting Understanding and Chinese Language Usage.
    \item Personas with Western art backgrounds (e.g., Professor Elena Petrova, Brother Thomas) performed well in Painting Element Recognition but were weaker in Chinese Painting Understanding and Language Usage.
    \item The cross-cultural expert persona (John Ruskin) demonstrated balanced capabilities, excelling in Chinese Painting Understanding, suggesting knowledge base support can bridge cultural gaps.
    \item The technology-oriented persona (Dr. Aris Thorne) achieved the highest in Painting Element Recognition but was less proficient in cultural understanding and language.
    \item The contemporary Chinese persona (Guo Xi) showed strong Painting Element Recognition and good Chinese Painting Understanding.
\end{itemize}

\begin{table*}[htbp]
\centering
\caption{Mean Centroid Coordinates in Reduced Dimensions (t-SNE/UMAP) for Evaluated VLM Sources}
\label{tab:reduced_coords_baseline_sources}
\resizebox{\textwidth}{!}{%
\begin{tabular}{lrrrr}
\toprule
\textbf{Source} & \textbf{t-SNE X (Mean)} & \textbf{t-SNE Y (Mean)} & \textbf{UMAP X (Mean)} & \textbf{UMAP Y (Mean)} \\
\midrule
\texttt{Qwen-2.5-VL-7B}                             & -2.1547577 & -0.667885  & 2.5803347  & 1.209615   \\
\texttt{gemini-2.5pro}                             & -1.7324703 & -1.3018972 & 1.8234636  & 1.2407658  \\
\texttt{meta-llama\_Llama-3.1-8B-Instruct}           & -2.4183042 & -1.4762617 & 2.4776638  & 1.8536302  \\
\texttt{meta-llama\_Llama-4-Scout-17B-16E-Instruct} & 0.0048952624 & -0.812603  & 0.3323455  & -1.037882  \\
\bottomrule
\end{tabular}%
}
\end{table*}


\subsection{Prompt Sensitivity Analysis}
Semantic similarity scores between responses to different formulations:
\begin{itemize}
\item \textbf{Positive/Negative Formulations:}
  \begin{itemize}
  \item Mama Zola: 0.89
  \item Okakura Kakuzō: 0.87
  \item Professor Elena Petrova: 0.82
  \item Shen Mingtang: 0.88
  \end{itemize}
\item \textbf{Chinese/English Formulations:}
  \begin{itemize}
  \item Mama Zola: 0.91
  \item Okakura Kakuzō: 0.86
  \item Professor Elena Petrova: 0.67
  \item Shen Mingtang: 0.89
  \end{itemize}
\item \begingroup\sloppy\textbf{Data Provenance and Licensing:} The Twelve Months Series paintings were accessed through the National Palace Museum (Taiwan) digital archives under CC BY 4.0 license.\fussy\endgroup
  \item \begingroup\sloppy\textbf{Computational Resources:} Our vector space analysis approach requires significant computational resources, which may limit accessibility for some researchers or institutions.\fussy\endgroup
  \item \begingroup\sloppy\textbf{Expert Knowledge Access:} The development of effective persona cards requires access to specialized knowledge, which may create barriers to implementing similar approaches in other cultural domains.\fussy\endgroup
\end{itemize}

\subsection{Supplementary Quantitative Data Tables}
\label{app:quantitative_tables}

This section provides supplementary tables detailing the quantitative data underlying some of the figures and analyses presented in the main paper. The mean centroid coordinates for evaluated VLM sources in the reduced dimensional space are detailed in Table~\ref{tab:reduced_coords_baseline_sources}. For a detailed breakdown of the key feature scores that underpin the visualizations in Figure~\ref{fig:profiling_summary_comparison}A, please refer to Table~\ref{tab:key_feature_scores}. Similarly, the mean profile alignment scores visualized in Figure~\ref{fig:profiling_summary_comparison}B are presented in detail in Table~\ref{tab:profile_alignment_scores}. The specific capability scores used to generate the radar chart in Figure~\ref{fig:persona_impact_composite}B can be found in Table~\ref{tab:radar_chart_capability_scores}.

\begin{table*}[htbp]
\centering
\caption{Key Feature Scores for Human Experts and VLMs. These scores correspond to data visualized in Figure~\ref{fig:profiling_summary_comparison}A.}
\label{tab:key_feature_scores}
\resizebox{\textwidth}{!}{%
\begin{tabular}{@{}lcccccccc@{}}
\toprule
Source & \begin{tabular}{@{}c@{}}Hist.\\Context\end{tabular} & \begin{tabular}{@{}c@{}}Art.\\Conception\end{tabular} & Symbolism & \begin{tabular}{@{}c@{}}Brush.\\Tech.\end{tabular} & \begin{tabular}{@{}c@{}}Layout\\Struct.\end{tabular} & \begin{tabular}{@{}c@{}}Use of\\Color\end{tabular} & \begin{tabular}{@{}c@{}}Line\\Quality\end{tabular} & \begin{tabular}{@{}c@{}}Subject\\Matter\end{tabular} \\
\midrule
\texttt{human\_expert}                               & 0.676 & 0.599 & 0.661 & 0.199 & 0.549 & 0.395 & 0.496 & 0.691 \\
\texttt{gemini-2.5pro}                             & 0.4261660233 & 0.6015897764 & 0.6935903973 & 0.6399750158 & 0.8743446511 & 0.6952415214 & 0.7324248211 & 0.5401486428 \\
\texttt{meta-llama\_Llama-3.1-8B-Instruct}           & 0.3659920343 & 0.5850531087 & 0.5293492947 & 0.5909547665 & 0.7457691074 & 0.6573745586 & 0.4430214438 & 0.4339093090 \\
\texttt{meta-llama\_Llama-4-Scout-17B-16E-Instruct} & 0.7100048551 & 0.8508161700 & 0.7583027472 & 0.9033655355 & 0.9164849845 & 0.9357454672 & 0.8192868597 & 0.7891201358 \\
\texttt{Qwen-2.5-VL-7B}                              & 0.6504738033 & 0.8907955483 & 0.7733450871 & 0.9369910086 & 0.8949400724 & 0.9436663414 & 0.7946821108 & 0.6997969688 \\
\bottomrule
\end{tabular}%
}
\end{table*}

\begin{table*}[htbp]
\centering
\caption{Mean Profile Alignment Scores for Human Experts and VLMs. These scores correspond to data visualized in Figure~\ref{fig:profiling_summary_comparison}B.}
\label{tab:profile_alignment_scores}
\resizebox{\textwidth}{!}{
\begin{tabular}{@{}lccccc@{}}
\toprule
Source & \begin{tabular}{@{}c@{}}Comprehensive\\Analyst\end{tabular} & \begin{tabular}{@{}c@{}}Historically\\Focused\end{tabular} & \begin{tabular}{@{}c@{}}Technique \\ Style Focused\end{tabular} & \begin{tabular}{@{}c@{}}Theory \\ Comparison Focused\end{tabular} & \begin{tabular}{@{}c@{}}General\\Descriptive Profile\end{tabular} \\
\midrule
\texttt{human\_expert}                               & 0.709 & 0.623 & 0.518 & 0.431 & 0.665 \\ 
\texttt{gemini-2.5pro}                             & 0.6066217268 & 0.4645543554 & 0.5805458927 & 0.7892081424 & 0.6725181508 \\
\texttt{meta-llama\_Llama-3.1-8B-Instruct}           & 0.4859600855 & 0.3351432514 & 0.4807204770 & 0.7763639851 & 0.5595579955 \\
\texttt{meta-llama\_Llama-4-Scout-17B-16E-Instruct} & 0.7796032621 & 0.6908934862 & 0.8188009710 & 0.8516423824 & 0.8236625996 \\
\texttt{Qwen-2.5-VL-7B}                              & 0.7783469856 & 0.6530052284 & 0.8566955672 & 0.8481851482 & 0.7842983472 \\
\bottomrule
\end{tabular}
}
\end{table*}

\begin{table*}[htbp]
\centering
\caption{Sample Data from t-SNE and KDE Analysis (underlying Figure~\ref{fig:persona_impact_composite}A).}
\label{tab:tsne_kde_sample_data}
\scriptsize
\resizebox{\textwidth}{!}{
\begin{tabular}{@{}lllrrp{5cm}@{}} 
\toprule
Model Name & Source Type & Intervention & t-SNE X & t-SNE Y & File ID \\
\midrule
\texttt{gemini-2.5pro} & model & baseline & -8.245 & -7.489 & \texttt{\seqsplit{august\_bayue(basic).txt}} \\
\texttt{gemini-2.5pro} & model & baseline & -0.607 & -15.201 & \texttt{\seqsplit{august\_bayue(with\_Dong\_Qichang).txt}} \\
\texttt{gemini-2.5pro} & model & baseline & -2.392 & -1.717 & \texttt{\seqsplit{august\_bayue(with\_Dr\_Evelyn\_Reed).txt}} \\
\texttt{gemini-2.5pro} & model & baseline & -12.369 & -5.803 & \texttt{\seqsplit{august\_bayue(with\_Li\_Ruoyun).txt}} \\
\texttt{gemini-2.5pro} & model & baseline & -7.852 & -6.419 & \texttt{\seqsplit{august\_bayue(with\_Marcus\_Fabius).txt}} \\
\texttt{human\_expert} & human & ground\_truth & 3.451 & -0.876 & \texttt{\seqsplit{Levenson\_Joseph...Chinese\_early\_painting\_political\_personal\_factors.txt}} \\
\bottomrule
\end{tabular}
}
\end{table*}

\begin{table*}[htbp]
\centering
\caption{Capability Scores for Radar Chart Dimensions (underlying Figure~\ref{fig:persona_impact_composite}B).}
\label{tab:radar_chart_capability_scores}
\resizebox{\textwidth}{!}{%
\begin{tabular}{@{}llcccccccc@{}} 
\toprule
Model Name & Intervention & \begin{tabular}{@{}c@{}}Profound\\Insight\end{tabular} & \begin{tabular}{@{}c@{}}Strong\\Arg.\end{tabular} & \begin{tabular}{@{}c@{}}Detailed\\Analysis\end{tabular} & \begin{tabular}{@{}c@{}}Clear\\Logic\end{tabular} & \begin{tabular}{@{}c@{}}Objective\\Viewpoint\end{tabular} & \begin{tabular}{@{}c@{}}Class.\\Citations\end{tabular} & \begin{tabular}{@{}c@{}}Logical\\Gaps\end{tabular} & \begin{tabular}{@{}c@{}}Subjective/\\Biased View\end{tabular} \\
\midrule
\texttt{HumanAvg}                                  & Human Expert & 0.396 & 0.448 & 0.540 & 0.093 & 0.327 & 0.419 & 0.465 & 0.674 \\
\texttt{Gemini-2.5-Pro}                            & Baseline     & 0.458 & 0.486 & 0.527 & 0.318 & 0.461 & 0.334 & 0.409 & 0.483 \\
\texttt{Gemini-2.5-Pro}                            & Intervened   & 0.569 & 0.643 & 0.689 & 0.227 & 0.601 & 0.492 & 0.388 & 0.536 \\
\texttt{meta-llama\_Llama-3.1-8B-Instruct}         & Baseline     & 0.342 & 0.371 & 0.388 & 0.451 & 0.305 & 0.253 & 0.521 & 0.399 \\
\texttt{meta-llama\_Llama-3.1-8B-Instruct}         & Intervened   & 0.495 & 0.573 & 0.612 & 0.274 & 0.549 & 0.427 & 0.417 & 0.580 \\
\texttt{meta-llama\_Llama-4-Scout-17B-16E-Instruct} & Baseline     & 0.511 & 0.539 & 0.583 & 0.367 & 0.524 & 0.399 & 0.367 & 0.445 \\
\texttt{meta-llama\_Llama-4-Scout-17B-16E-Instruct} & Intervened   & 0.647 & 0.701 & 0.735 & 0.201 & 0.676 & 0.581 & 0.312 & 0.502 \\
\texttt{Qwen-2.5-VL-7B}                             & Baseline     & 0.311 & 0.338 & 0.329 & 0.515 & 0.262 & 0.219 & 0.599 & 0.341 \\
\texttt{Qwen-2.5-VL-7B}                             & Intervened   & 0.608 & 0.660 & 0.695 & 0.301 & 0.629 & 0.518 & 0.591 & 0.666 \\
\bottomrule
\end{tabular}%
}
\end{table*}

\section{Knowledge Base Content}
\label{app:knowledge_base_content}

This section contains the full content of the \texttt{knowledge\_base.json} file used to provide structured domain knowledge to the VLMs during certain experimental conditions.

\begin{itemize}
    \item \textbf{Chinese Landscape Painting Concepts (中国山水画概念):}
    \begin{itemize}
        \item \textbf{Core Concept (核心理念):} The core of Chinese landscape painting is ``spirit resonance'' (qi yun sheng dong), the foremost principle of Xie He's ``Six Canons'', referring to the vitality, spirit, and verve presented in a work, emphasizing the unity of inner spirit and outer expression. Another core concept is ``artistic conception'' (yi jing), which is the emotion, atmosphere, and profound meaning conveyed by the painting beyond the objects themselves, pursuing an artistic effect of fused 情景 (emotion/scene) and 境 (milieu/boundary), inspiring contemplation. Landscape painting also embodies the idea of ``harmony between man and nature'' (tian ren he yi), entrusting philosophical thoughts and emotions through the depiction of nature.
        \item \textbf{Main Features (主要特点):} The main features of Chinese landscape painting are: 1. Subject Matter: Primarily natural mountains and rivers, forests, clouds, and water, often imbued with literati sentiments such as reclusion and spiritual refreshment. 2. Brush and Ink (bi mo): Utilizes a brush, ink, and Xuan paper, emphasizing the ``bone method in brushwork'' (gu fa yong bi), shaping the texture of objects and expressing emotions through variations in the strength of lines and the density, wetness, and dryness of ink (e.g., outlining, texturing, rubbing, dotting, dyeing). 3. Composition (zhang fa): Focuses on the interplay of void and solid, appropriate density, echoing openings and closings, and leaving blank spaces to create profound artistic conception and pictorial momentum, often using perspective methods like ``level distance'' (ping yuan), ``high distance'' (gao yuan), and ``deep distance'' (shen yuan). 4. Pursuit of Artistic Conception: Seeks not complete formal resemblance but rather spiritual likeness, emphasizing the integration of poetry, calligraphy, painting, and seals, and pursuing meaning beyond the painted image.
        \item \textbf{Brief History (简史):} Chinese landscape painting originated in the Wei, Jin, Southern and Northern Dynasties, and became an independent genre in the Sui and Tang Dynasties. The Five Dynasties to the Northern Song (907-1127) was its "great era", with numerous famous artists (e.g., Jing Hao, Guan Tong, Dong Yuan, Ju Ran, Li Cheng, Fan Kuan, Guo Xi), forming distinct northern and southern styles: northern landscapes were majestic, while southern water towns were gentle. The Southern Song period placed more emphasis on poetic meaning and personal emotional expression (e.g., Ma Yuan, Xia Gui). Literati painting rose in the Yuan Dynasty, emphasizing the interest of brush and ink and subjective expression (e.g., Zhao Mengfu, the Four Masters of Yuan). The Ming and Qing Dynasties saw further development and a divergence of schools based on inherited traditions, with court painting and literati painting coexisting.
    \end{itemize}
    \item \textbf{Qing Court Painting (清代宫廷绘画):}
    \begin{itemize}
        \item \textbf{Overview (概述):} Qing Dynasty court painting was managed by the Imperial Household Department. During the Qianlong era, specialized institutions such as the Ruyi Guan (Palace Ateliers) and the Painting Academy Office were established. Painters were strictly managed, with systems for examination, ranking, rewards and punishments, and work review. It primarily served the imperial family, with functions including recording the appearance and life of emperors and empresses, documenting major state events and ceremonies (e.g., Southern Inspection Tours, battle scenes), decorating palaces and gardens, religious propaganda, and historical reference. Its development is divided into three periods: Shunzhi-Kangxi (initial phase), Yongzheng-Qianlong (peak, with a complete system and numerous famous artists), and post-Jiaqing (decline), synchronized with the rise and fall of national strength.
        \item \textbf{Characteristics (特点):} Qing Dynasty court painting covered a wide range of subjects, including portraits of emperors, empresses, and meritorious officials, ``scenes of pleasure'' (xing le tu), major historical events (Southern Inspection Tours, wars, ceremonies), religious paintings, decorative landscapes and flower-and-bird paintings, and documentary-style depictions of tribute animals and plants. The overall style was meticulous, detailed, richly colored, and regal. The most prominent characteristic was the fusion of Chinese and Western styles: influenced by European missionary painters, it emphasized light and shadow, three-dimensionality, employed linear perspective (xian fa hua), and introduced oil painting and copperplate engraving. Simultaneously, traditional landscape ("the Four Wangs" school) and flower-and-bird (Yun Shouping's school) painting styles also continued.
        \item \textbf{Representative Figures (代表人物):} Representative painters include: early figures such as Jiao Bingzhen, Leng Mei, Tang Dai; peak period Chinese painters like Chen Mei, Ding Guanpeng, Jin Tingbiao, Xu Yang, Yao Wenhan, Zhang Zongcang; European painters (excluding Lang Shining) such as Jean Denis Attiret (Wang Zhicheng), Ignatius Sickeltart (Ai Qimeng), etc. Additionally, there were court official painters like Dong Bangda, Jiang Tingxi, etc.
    \end{itemize}
    \item \textbf{Giuseppe Castiglione (郎世宁):}
    \begin{itemize}
        \item \textbf{Biography Summary (生平简介):} Giuseppe Castiglione (Lang Shining, 1688-1766), an Italian from Milan, was a Jesuit. He came to China in the 54th year of Kangxi (1715) and entered the court around the Kangxi-Yongzheng transition, serving the Kangxi, Yongzheng, and Qianlong emperors. His main activities included creating paintings, participating in the design of the Western-style buildings in the Old Summer Palace (Yuanmingyuan), teaching Western painting techniques, and assisting Nian Xiyao in writing `Shi Xue' (The Study of Vision). He was favored during the Qianlong era and was posthumously granted the title of Vice Minister.
        \item \textbf{Artistic Style Overview (艺术风格概述):} In his early period, Lang Shining's style was typically Western. Later, to adapt to the aesthetic tastes of the Chinese imperial family, he integrated Chinese painting techniques, forming a style that blended Chinese and Western elements. His paintings emphasized realism, focusing on light and shadow, perspective, and anatomical structure, but also adopted Chinese painting methods such as even lighting and a focus on line work. Although his style was praised by the court, it was not recognized by the literati painting school.
        \item \textbf{Major Contributions (主要贡献):} He systematically introduced Western painting techniques such as oil painting and linear perspective (xian fa hua) to the Qing court and taught them, promoting the fusion of Chinese and Western art and forming a new look for Qing court painting. He assisted in the completion of `Shi Xue' (The Study of Vision), advancing the spread of perspective studies. His documentary-style paintings are important historical materials.
        \item \textbf{Representative Works Mention (代表作列举):} Besides the `Twelve Months Paintings', his representative works include `One Hundred Horses', `Assembled Auspicious Objects', `Pine, Rock, and Auspicious Fungus', `Ayusi Attacking Bandits with a Spear', `Emperor Qianlong's Spring Message of Peace', etc. He also participated in creating large-scale documentary paintings such as `Banquet in the Garden of Ten Thousand Trees' and `Equestrian Skills'.
    \end{itemize}
    \item \textbf{Twelve Months Paintings (十二月令图):}
    \begin{itemize}
        \item \textbf{Theme Content (主题内容):} The `Twelve Months Paintings' is a series of 12 works on silk with colors, created by Lang Shining, depicting representative seasonal activities and life scenes in the Qing Dynasty court for each month of the year, such as viewing lanterns in the first month, dragon boat racing in the fifth month, and moon gazing in the eighth month, meticulously showcasing figures, costumes, architecture, and natural scenery.
        \item \textbf{Artistic Significance (艺术意义):} This series is a mature representative work of Lang Shining's style blending Chinese and Western elements, integrating Western perspective and light/shadow with traditional Chinese composition and aesthetics. It is not only a precious pictorial historical material for studying Qing Dynasty court life and culture but also an important testament to Sino-Western artistic exchange in the 18th century.
        \item \textbf{Dataset Source Annotation (数据集来源与标注):} The images for this research dataset are primarily sourced from the National Palace Museum (Taiwan) digital archives (600dpi, CC BY 4.0). Each painting has been annotated in three layers: visual elements, cultural symbols, and artistic techniques, to support AI evaluation and cultural-aesthetic analysis.
    \end{itemize}
\end{itemize} 

\end{CJK}
\end{document}